\documentclass[runningheads]{llncs}
\usepackage{graphicx}

\usepackage{tikz}
\usepackage{comment}
\usepackage{amsmath,amssymb} 
\usepackage{color}

\usepackage[accsupp]{axessibility}  

\usepackage[width=122mm,left=12mm,paperwidth=146mm,height=193mm,top=12mm,paperheight=217mm]{geometry}

\usepackage[pagebackref=true,breaklinks=true,colorlinks,citecolor=gray]{hyperref}
\usepackage{colortbl}
\usepackage{tabularx}
\usepackage{graphicx}
\usepackage{booktabs}
\usepackage{multirow}
\usepackage{amsfonts,bm}
\usepackage{algorithm}
\usepackage{algorithmic}
\usepackage{xspace}

\newcommand{\tabincell}[2]{\begin{tabular}{@{}#1@{}}#2\end{tabular}}

\newcolumntype{x}[1]{>{\centering\arraybackslash\hspace{0pt}}p{#1}}

\definecolor{Gray}{gray}{0.93}
\definecolor{orange}{rgb}{0.9,0.5,0}
\usepackage{amsmath}
\usepackage{upgreek}
\usepackage{amssymb}
\usepackage{pifont}%
\usepackage{multirow}
\usepackage{makecell}

\newcommand{\ie}{\emph{i.e.}}
\newcommand{\eg}{\emph{e.g.}}
\newcommand{\model}{DaViT\xspace}
\newcommand{\modelInTable}{DaViT\xspace}
\newcommand{\modelFull}{Dual Attention Vision Transformers\xspace}
\newlength\savewidth
\newcommand\shline{\noalign{\global\savewidth\arrayrulewidth
  \global\arrayrulewidth 1pt}\hline\noalign{\global\arrayrulewidth\savewidth}}

\begin{document}
\pagestyle{headings}
\mainmatter
\def\ECCVSubNumber{812}  

\title{\model: Dual Attention Vision Transformers} 

\titlerunning{\model: Dual Attention Vision Transformers}

\author{
\hspace{-14pt}Mingyu Ding$^{1}$,
Bin Xiao$^{2}$\thanks{Corresponding author},
Noel Codella$^{2}$,
Ping Luo$^{1\star}$,
Jingdong Wang$^{3}$,
Lu Yuan$^{2}$
\\
$^{1}$The University of Hong Kong \quad 
$^{2}$Microsoft Cloud + AI \quad
$^{3}$Baidu \\
{
\tt\small mingyuding@hku.hk \quad 
\tt\small \{bixi,ncodella,luyuan\}@microsoft.com \\
\tt\small pluo@cs.hku.hk \quad 
\tt\small wangjingdong@outlook.com
}
}
\institute{}

\authorrunning{M. Ding et al.}
%
\maketitle

\begin{abstract}
In this work, we introduce \modelFull~(\model), a simple yet effective vision transformer architecture that is able to capture global context while maintaining computational efficiency. We propose approaching the problem from an orthogonal angle: exploiting self-attention mechanisms with both {\em ``spatial tokens''} and ``{\em channel tokens}''. With spatial tokens, the spatial dimension defines the token scope, and the channel dimension defines the token feature dimension. With channel tokens, we have the inverse: the channel dimension defines the token scope, and the spatial dimension defines the token feature dimension. 
We further group tokens along the sequence direction for both spatial and channel tokens to maintain the linear complexity of the entire model.
We show that these two self-attentions complement each other: 
(i) since each channel token contains an abstract representation of the entire image, the channel attention naturally captures global interactions and representations by taking all spatial positions into account when computing attention scores between channels;
(ii) the spatial attention refines the local representations by performing fine-grained interactions across spatial locations, which in turn helps the global information modeling in channel attention.
Extensive experiments show our \model achieves state-of-the-art performance on four different tasks with efficient computations.
Without extra data, \modelInTable-Tiny, \modelInTable-Small, and \modelInTable-Base achieve 82.8\%, 84.2\%, and 84.6\% top-1 accuracy on ImageNet-1K with 28.3M, 49.7M, and 87.9M parameters, respectively. When we further scale up \model with 1.5B weakly supervised image and text pairs, \modelInTable-Gaint reaches 90.4\% top-1 accuracy on ImageNet-1K.
Code is available at
\href{https://github.com/dingmyu/davit}{\tt https://github.com/dingmyu/davit}.
\end{abstract}

\section{Introduction}

The global context is essential for many computer vision approaches, such as image classification and semantic segmentation.
Convolutional neural networks (CNNs)~\cite{liu2022convnet} gradually obtain a global receptive field by multi-layer architectures and down-sampling operators.
Recently, vision transformers~\cite{chen2020generative,dosovitskiy2020image}, which directly capture long-range visual dependencies with a single self-attention layer, have drawn much attention.
While these methods present strong capabilities to model the global context, their computational complexity grows quadratically with the token length, limiting its ability to scale up to high-resolution scenarios.

\begin{figure}[t]
    \centering
    \includegraphics[width=0.99\linewidth]{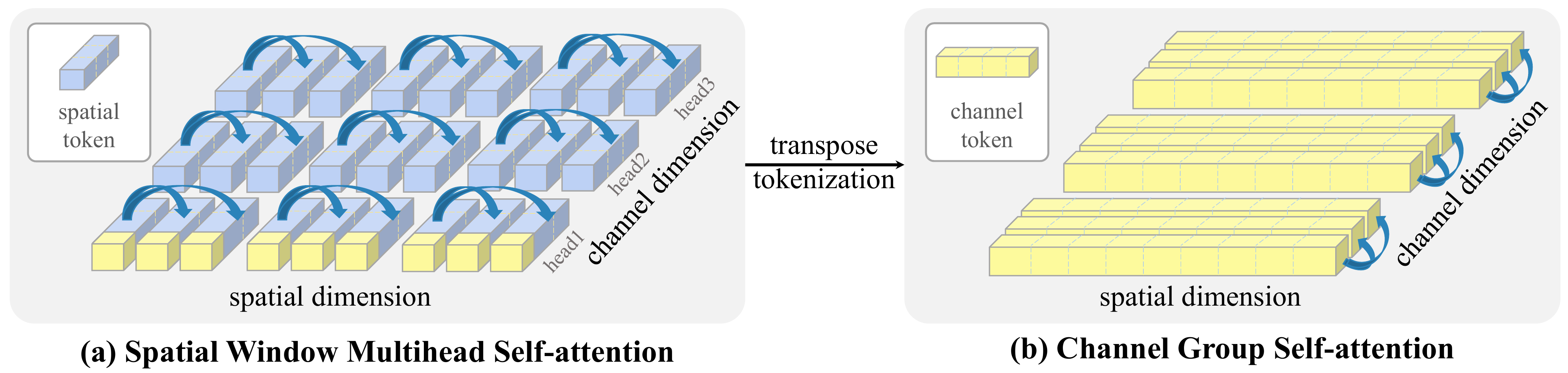}
    \vspace{-8pt}
    \caption{(a) Spatial window multihead self-attention splits the spatial dimension into local windows, where each window contains multiple spatial tokens. Each token is also divided into multiple heads.
    (b) Channel group single-head self-attention groups channel tokens into multi groups. Attention is performed in each channel group with an entire image-level channel as a token.
    A channel-wise token that captures global information is also highlighted in (a).
    In this work, we alternately use these two types of attention to obtain both local fine-grained and global features.
    }
    \label{fig:teaser}
    \vspace{-8pt}
\end{figure}

Designing an architecture that can capture global contexts while maintaining efficiency to learn from high-resolution inputs is still an open research problem. 
A substantial body of work has been dedicated to developing vision transformers toward this goal.
iGPT~\cite{chen2020generative} first utilized a standard transformer to solve vision tasks by treating the image as sequences of pixels and performing pixel-level interactions.
After that, ViT~\cite{dosovitskiy2020image} used non-overlapped image patches as tokens to model the relationship between small image patches instead of pixels, showing promising performance on middle-resolution tasks such as classification.
To further reduce the computational cost, local attention~\cite{liu2021swin,zhang2021multi,vaswani2021scaling} that limits attention in a spatially local window, and squeezed projection~\cite{wu2021cvt,wang2021pvtv2} that performs attention on downsampled tokens, were proposed.
Though local attention methods benefit from linear complexity with the spatial size, operators like ``Shift''~\cite{liu2021swin}, ``Overlapping Patch''~\cite{wu2021cvt,wang2021pvtv2}, ``ConvFFN''~\cite{wang2021pvtv2,yuan2021hrformer,xie2021segformer} are indispensable to compensate for the loss of global contextual information.

The general pattern across all prior works is that they attain various tradeoffs between resolution, global context, and computational complexity: pixel-level~\cite{chen2020generative} and patch-level~\cite{dosovitskiy2020image,liu2021swin,wu2021cvt,wang2021pyramid} self-attentions suffer either the cost of quadratic computational overhead or loss of global contextual information.
Beyond variations of pixel-level and patch-level self-attentions, can we design an image-level self-attention mechanism that captures global information but is still efficient concerning the spatial size?

In this work, we introduce such a self-attention mechanism that is able to capture global context while maintaining computational efficiency.
In addition to {\em ``spatial tokens''} defined by existing works in Fig~\ref{fig:teaser}(a) representing the feature of an image patch, we introduce ``{\em channel tokens}'' by applying self-attention to the {\em transpose} of the token matrix, as shown in Fig~\ref{fig:teaser}(b). With channel tokens, the channel dimension defines the token scope, and the spatial dimension defines the token feature dimension.
In this way, each channel token is global on the spatial dimension, containing an abstract representation of the entire image. Correspondingly, performing self-attention on such channel tokens further captures the global interaction by taking all spatial positions into account when computing attention scores between channels.
Compared to conventional self-attention that performs global interactions over local pixels or patch tokens at a quadratic computational cost, the information exchange of channel self-attention is naturally imposed from a global perspective rather than a pixel/patch-wise one.
Based on the global receptive field of the channel token, it fuses the representations to produce new global tokens and passes the information to the following layers.
Thus, one can take such channel self-attention as a dynamic feature fusion over a series of abstract representations of the entire image.

\begin{figure}[t]
  \centering
        \includegraphics[width=0.24\linewidth]{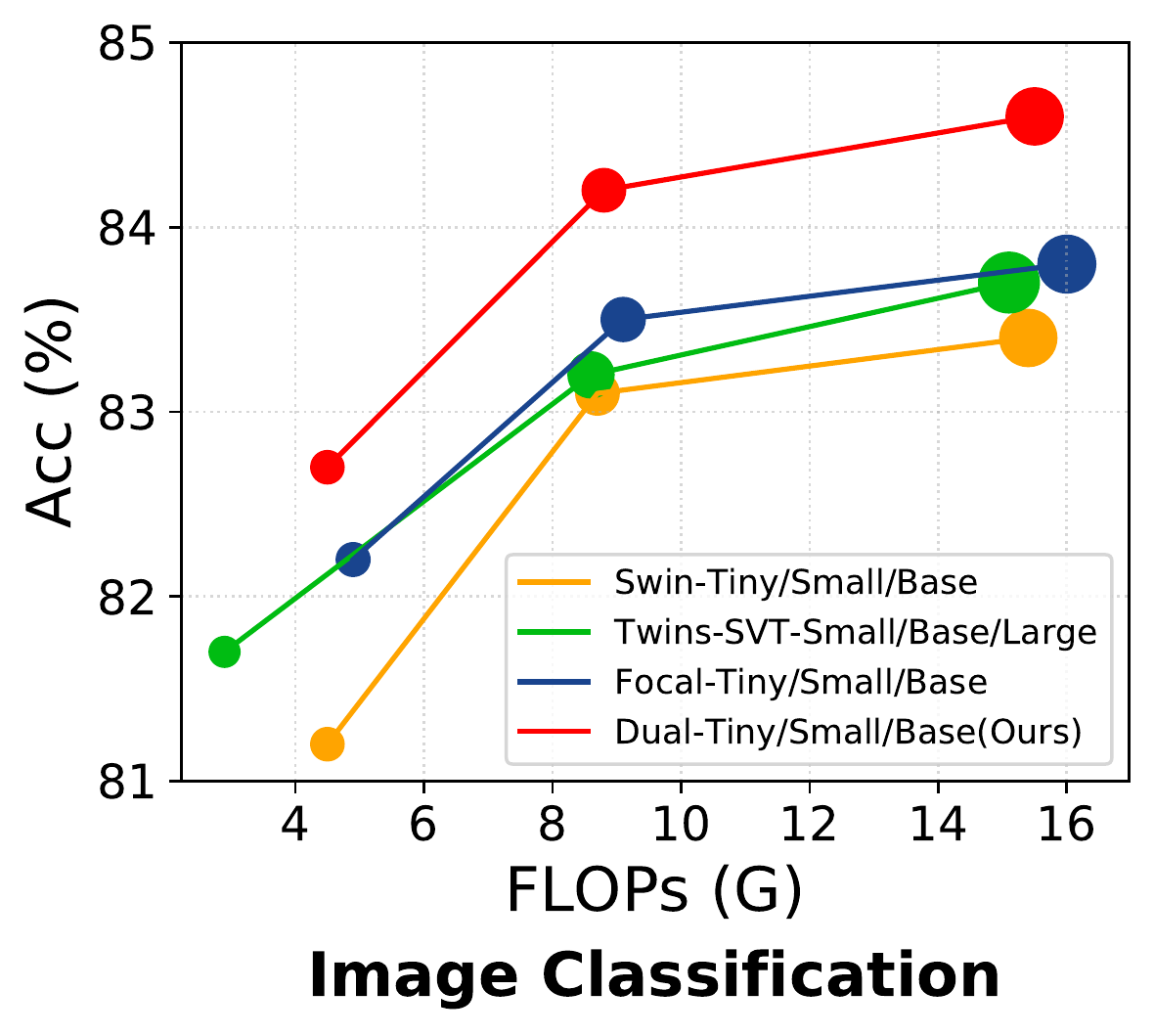}
        \includegraphics[width=0.24\linewidth]{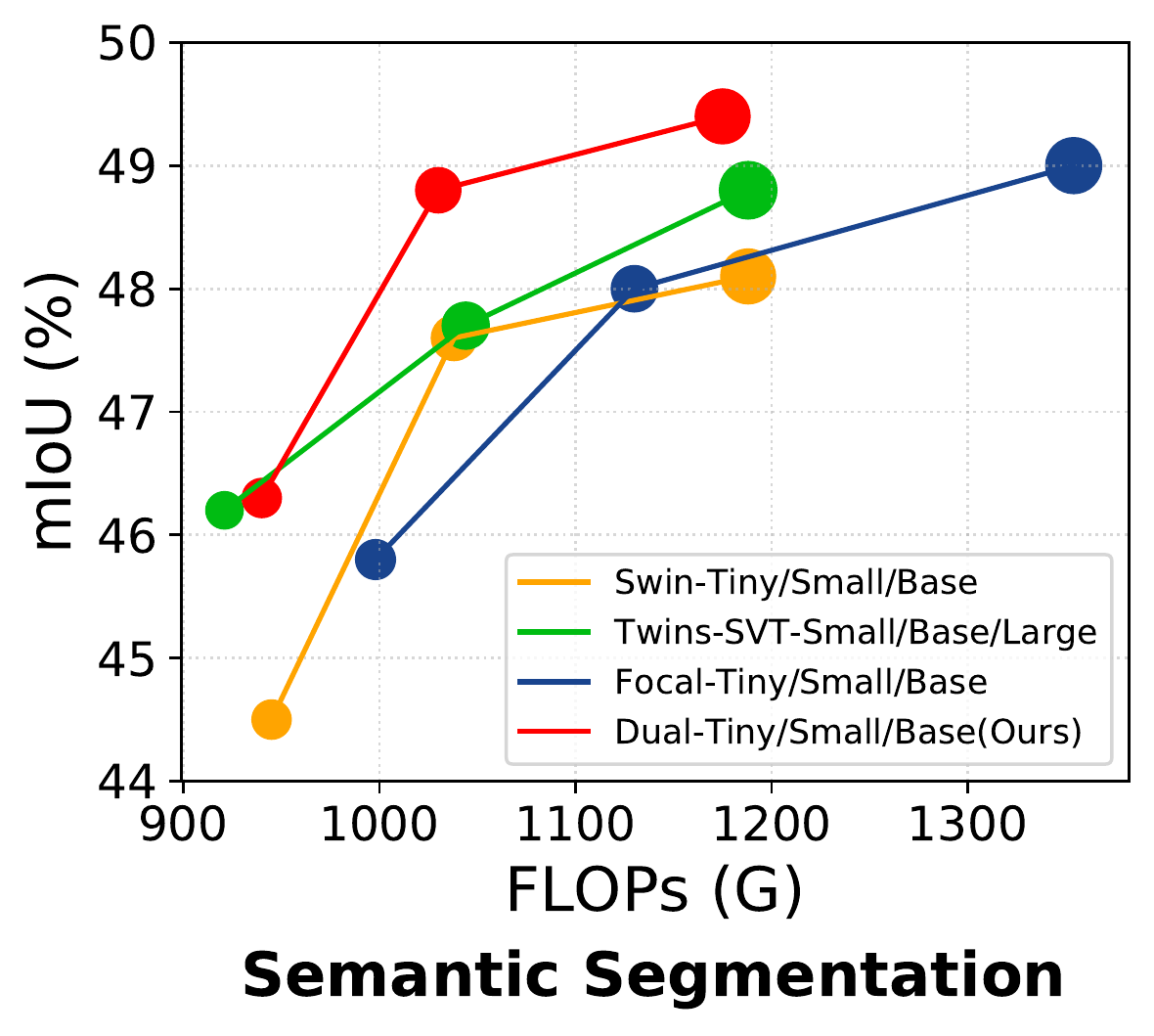}
        \includegraphics[width=0.24\linewidth]{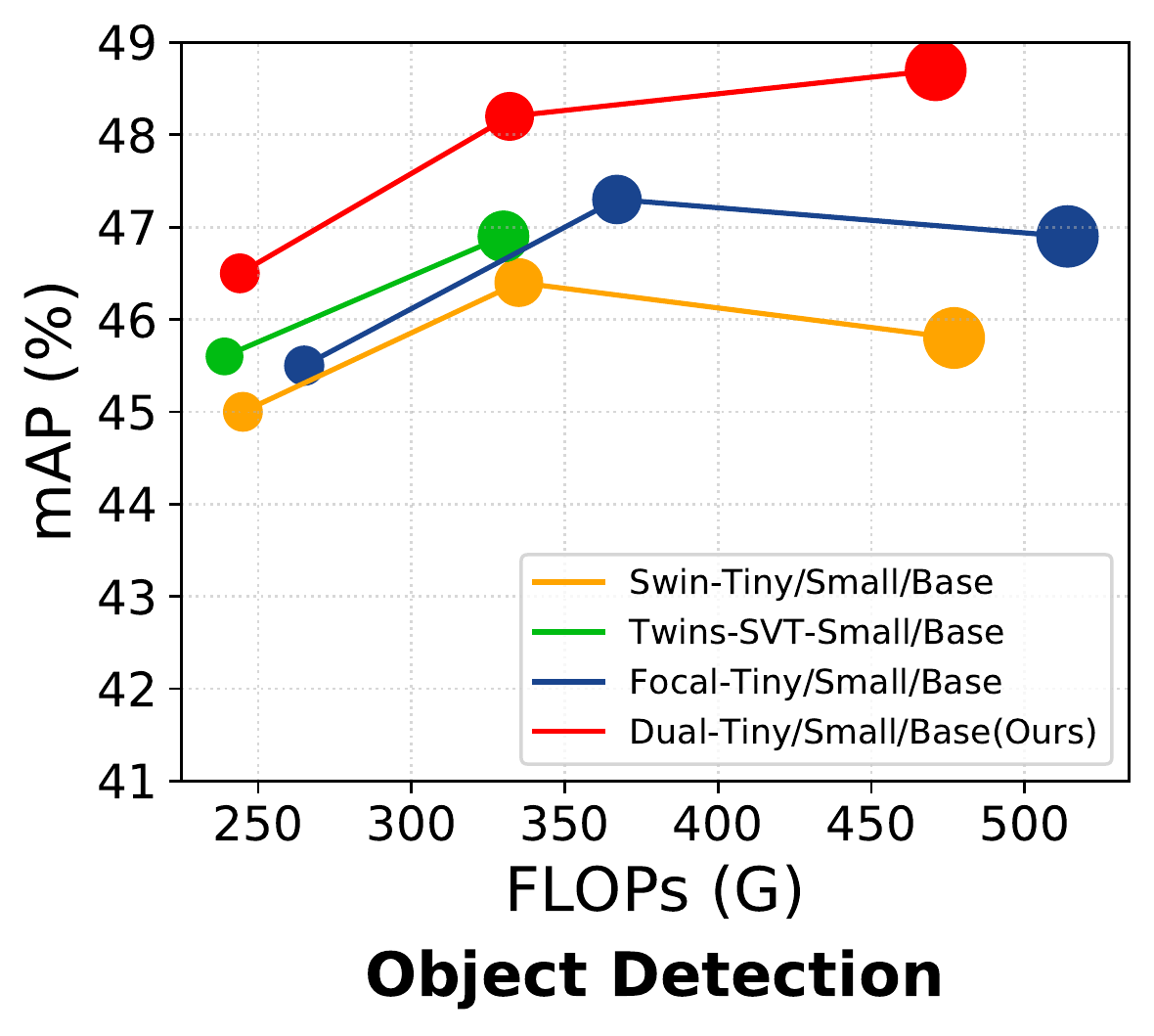}
        \includegraphics[width=0.24\linewidth]{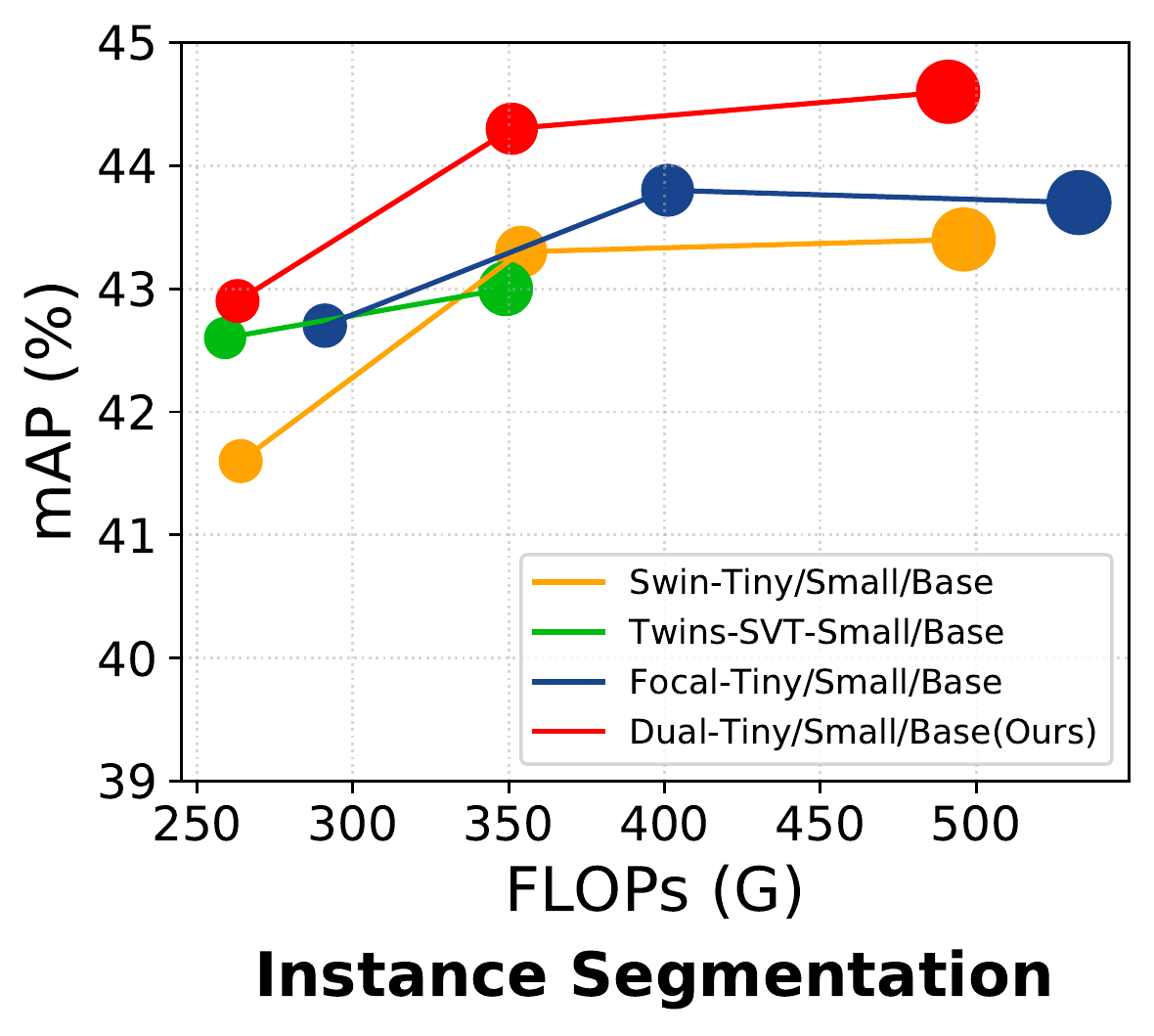}
  \vspace{-8pt}
  \caption{Comparisons of the efficiency (\ie, FLOPs) and performance (\eg, Acc, mIoU, mAP) between the proposed approach and existing SoTA methods~\cite{liu2021swin,yang2021focal,chu2021twins} on four computer vision tasks. Each method is represented by a circle, whose size represents the number of parameters. Our approach achieves superior performance under similar FLOPs than its counterparts on all four benchmarks.
  }
  \label{fig:comparison}
  \vspace{-8pt}
\end{figure}

Although this approach presents many advantages, a few challenges must be overcome. 
First, the computational complexity suddenly increases quadratically with the channel dimension, limiting the representation power of the layers. Inspired by spatial local attention~\cite{liu2021swin,zhang2021multi,vaswani2021scaling}, we propose {\em channel group attention} by dividing the feature channels into several groups and performing image-level interactions within each group. By group attention, we reduce the complexity to linear with respect to both the spatial and the channel dimensions. 
Since each channel token contains an abstract representation of the entire image, self-attention in this setting naturally captures the global interaction even we apply it locally along the channel dimensions.

Second, though channel-wise self-attention can capture global information easily, image-level tokens hinder local interactions across spatial locations.
To solve this problem, we introduce \modelFull~(\model) that alternately applies spatial window attention and channel group attention to capture both short-range and long-range visual dependencies, as shown in Figure~\ref{fig:teaser}.
Our results show that these two structures complement each other: the channel group attention provides a global receptive field on the spatial dimension and extracts high-level global-image representations by dynamic feature fusion across global channel tokens; the spatial window attention refines the local representations by performing fine-grained local interactions across spatial locations, which in turn helps the global information modeling in channel attention.

To summarize, we propose \model that contains two seamlessly integrated self-attentions: spatial window self-attention with {\em ``spatial tokens''}, and channel group self-attention with {\em ``channel tokens''}. The two forms of attention are complementary and alternatively arranged, providing both local fine-grained and global interactions in a computationally efficient manner.
We evaluate the effectiveness of the proposed \model via comprehensive empirical studies on image classification, object detection, and segmentation. Results in Figure~\ref{fig:comparison} show that \model consistently outperforms the SoTA vision transformers across three benchmarks and four tasks with even fewer computational costs.

\section{Related Work}

\noindent \textbf{Vision Transformers and MLPs.}~~
Transformers~\cite{vaswani2017attention,devlin2019bert} have dominated a wide range of natural language processing tasks. 
In computer vision, pioneering works iGPT~\cite{chen2020generative} and ViT~\cite{dosovitskiy2020image} apply attention directly to a sequence of image pixels or patches.
Similarly, follow-up works~\cite{heo2021rethinking,touvron2021training,wang2021pyramid,srinivas2021bottleneck,wu2021rethinking,graham2021levit,XiaohuaZhai2021ScalingVT,riquelme2021scaling,MichaelSRyoo2021TokenLearnerWC} model global relationships on the patch-level tokens.
All these works apply attention to capture interactions over all the local tokens, which can be pixel-level or patch-level.
In this work, we model the global relationship from an orthogonal perspective and apply attention mechanisms to both spatial tokens as well as their transpose, which we refer to as image-level (global) channel tokens. In this manner, we capture both fine-grained structural patterns and global interactions.

Most recently, MLP-based models~\cite{mlp-mixer,resmlp,gmlp,yu2022s2,chen2021cyclemlp,lian2021mlp,hou2022vision} draw attention by using multi-layer perceptrons (MLPs) directly over patch features.
MLP-Mixer~\cite{mlp-mixer} leverages feature transpose by applying MLPs across either spatial locations or feature channels repeatedly. However, it resorts to neither transformer architectures nor self-attention layers, and works only on the image classification task. Our work focuses on transformer-based architectures to benefit various tasks, \eg, classification, objection detection, and segmentation.

\noindent \textbf{Hierarchical Vision Transformers.}~~
Hierarchical designs are widely adopted to transformers~\cite{liu2021swin,wang2021pyramid,wu2021cvt,ding2021hr,zhang2021multi,vaswani2021scaling,pan2021scalable,yuan2021volo,li2021localvit,ali2021xcit,li2022uniformer,zhou2021deepvit,tang2022quadtree,vaswani2021scaling,chen2021visformer,li2021bossnas,yu2021glance,huang2021shuffle,xu2021vitae,YanghaoLi2021ImprovedMV,JingkaiZhou2021ELSAEL} in vision.
PVT~\cite{wang2021pyramid} and CvT~\cite{wu2021cvt} perform attention on the squeezed tokens to reduce the computational cost.
Swin Transformer~\cite{liu2021swin}, ViL~\cite{zhang2021multi}, and HaloNet~\cite{vaswani2021scaling} apply local windows attention to the patch tokens, which capture fine-grained features and reduce the quadratic complexity to linear, but lose the ability of global modeling. To compensate for the loss of global context, Swin Transformer~\cite{liu2021swin} conducts attention on the shifted local windows alternatively between consecutive blocks, and ViL~\cite{zhang2021multi} and HaloNet~\cite{vaswani2021scaling} play on overlapped windows.

In this work, the proposed approach shares merits of hierarchical architectures and fine-grained local attention, meanwhile our proposed group channel attention still efficiently models the global context.

\noindent \textbf{Channel-wise Attentions.}~~
The author of~\cite{hu2018squeeze} first proposes a Squeeze-and-Excitation (SE) block as channel-wise attention to re-calibrating the channel-wise features through the squeezed global feature. Other operators in CNNs related to our work are Dynamic Head~\cite{dai2021dynamic} and DANet~\cite{fu2019dual}. They apply attention along different feature dimensions on top of the CNN backbone for a specific task. 
Some transformer architectures involve channel-wise operations as well to reduce the computational costs.
LambdaNetworks~\cite{bello2021lambdanetworks} first transforms the context into a linear function lambda that is applied to the corresponding query.
XCiT~\cite{ali2021xcit} proposes cross-covariance attention (XCA) for efficient processing of high-resolution images.
Similarly, CoaT~\cite{xu2021co} introduces a factorized attention mechanism that works efficiently in a multi-branch transformer backbone.

Our work proposes channel group attention to capture global information in transformers. Though it has only linear complexity with respect to both channel and spatial dimensions, we demonstrate its power when combined with spatial window attention, forming our dual attention mechanism.
Furthermore, we analyze in detail how our dual attention obtains global interactions as well as fine-grained local features, showing its effectiveness in benefiting various tasks, \eg, classification, objection detection, and segmentation.

\begin{figure*}[t]
    \centering
    \includegraphics[width=0.99\textwidth]{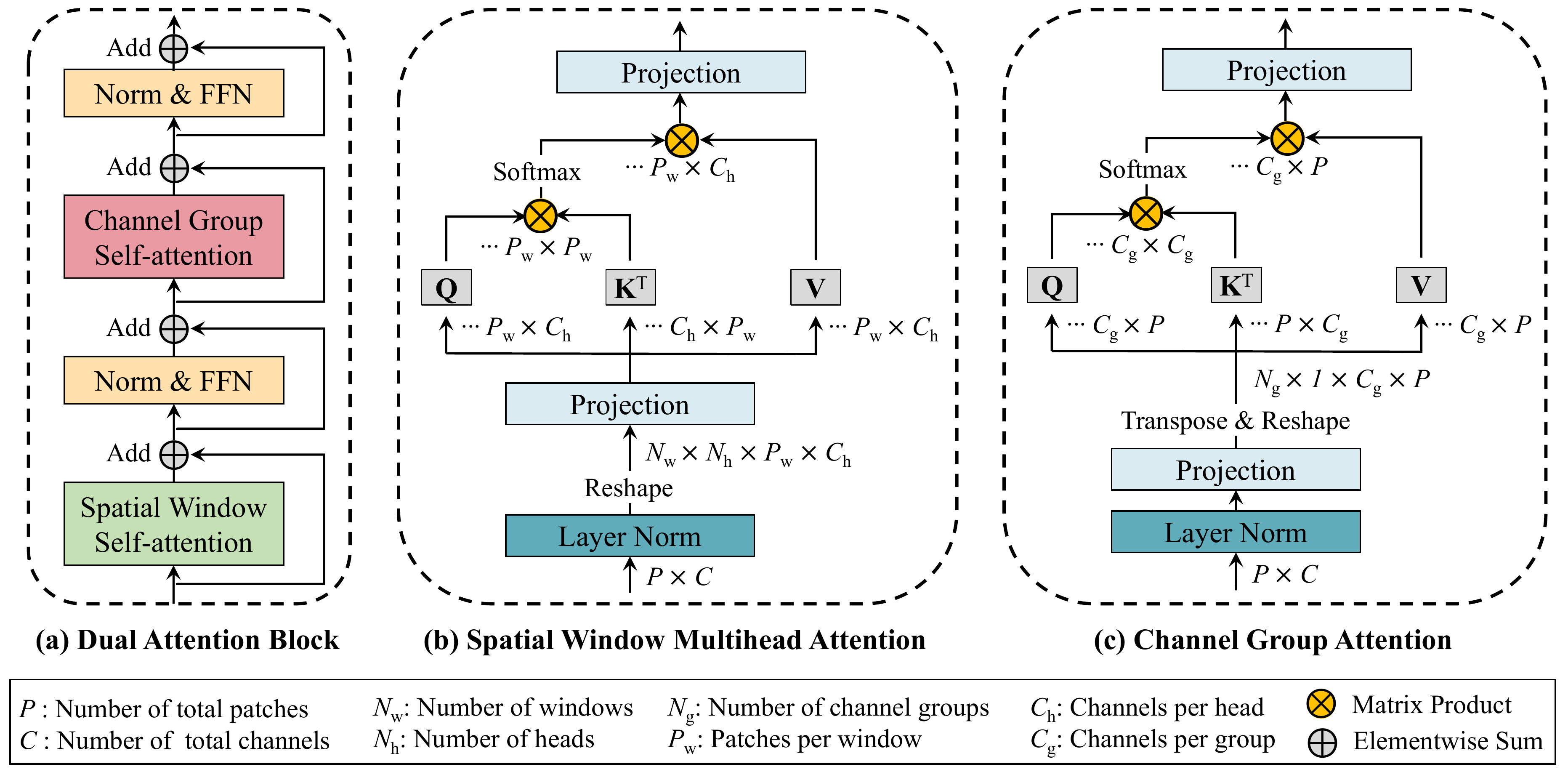}
    \vspace{-8pt}
    \caption{Model architecture for our dual attention block.
    It contains two transformer blocks: spatial window self-attention and channel group self-attention blocks.
    By alternately using the two types of attention, our model enjoys the benefit of capturing both local fine-grained and global image-level interactions.
    }
    \label{fig:architecture}
    \vspace{-8pt}
\end{figure*}

\section{Methodology}
We propose \modelFull (\model), a clean, efficient, yet effective transformer backbone containing both local fine-grained features and global representations.
In this section, we first introduce the hierarchical layout of our model. We then detail our channel group attention and the combination with spatial window attention~\cite{liu2021swin}.

\subsection{Overview}
We divide the model into four stages, where a patch embedding layer is inserted at the beginning of each stage. We stack our dual attention blocks in each stage with the resolution and feature dimension kept the same. Figure~\ref{fig:architecture}(a) illustrates the architecture of our dual attention block, consisting of a spatial window attention block and a channel group attention block.

\noindent \textbf{Preliminaries.}~~
Let us assume a $\mathbb{R}^{P \times C}$ dimensional visual feature, 
where $P$ is the number of total patches and $C$ is the number of total channels. 
Simply applying the standard global self-attention leads to a complexity of $O(2P^2C + 4PC^2)$.
It is defined as:
\begin{align}
\mathcal{A}(\mathbf{Q}, \mathbf{K}, \mathbf{V}) & = \mathrm{Concat}(\mbox{head}_1,\ldots,\mbox{head}_{N_h}) \notag \\
\text{where}~~\mbox{head}_i & = \mathrm{Attention}(\mathbf{Q}_i, \mathbf{K}_i, \mathbf{V}_i) \notag \\
& = \mathrm{softmax} \left[\frac{\mathbf{Q}_i(\mathbf{K}_i)^\mathrm{T}}{\sqrt{C_h}}\right]\mathbf{V}_i
\label{eq:self-attention}
\end{align}
where $\mathbf{Q}_i=\mathbf{X}_i\mathbf{W}_i^Q$, $\mathbf{K}_i=\mathbf{X}_i\mathbf{W}_i^K$, and $\mathbf{V}_i=\mathbf{X}_i\mathbf{W}_i^V$ are $\mathbb{R}^{P \times C_h}$ dimensional visual features with $N_h$ heads, $\mathbf{X}_i$ denotes the $i_{th}$ head of the input feature and $\mathbf{W}_i$ denotes the projection weights of the $i_{th}$ head for $\mathbf{Q}, \mathbf{K}, \mathbf{V}$, and $C = C_h * N_h$.
Please note that the output projection $\mathbf{W}^O$ is omitted here.
Considering $P$ can be very large, \eg, $128 \times 128$, the computational cost is immoderate.

In \model, we alternatively arrange spatial window attention and channel group attention to obtain both local and global features, but with a linear complexity to the spatial dimension, as shown in Figure~\ref{fig:architecture}.

\subsection{Spatial Window Attention}
Window attention computes self-attention within local windows, as shown in Figure~\ref{fig:teaser}(a). The windows are arranged to partition the image in a non-overlapping manner evenly. Supposing there are $N_w$ different windows with each window containing $P_w$ patches, where $P = P_w * N_w$.
Then window attention can be represented by:
\begin{equation}
 \mathcal{A}_{window}(\mathbf{Q}, \mathbf{K}, \mathbf{V}) =  
 \{\mathcal{A}(\mathbf{Q}_i, \mathbf{K}_i, \mathbf{V}_i)\}_{i=0}^{N_w}
\label{eq:window-attention}
\end{equation}
where $\mathbf{Q}_i, \mathbf{K}_i, \mathbf{V}_i \in \mathbb{R}^{P_w \times C_h}$ are local window queries, keys, and values.
The computational complexity of a window-based self-attention is $O(2PP_wC + 4PC^2)$ with a linear complexity with the spatial size $P$. More details of window attention are shown in Figure~\ref{fig:architecture}(b).

Though the computation is reduced, window attention loses the ability to model the global information.
We will show that our proposed channel attention naturally solves this problem and mutually benefits with window attention.

\subsection{Channel Group Attention}
We visit self-attention from another perspective and propose channel-wise attention, as shown in Figure~\ref{fig:architecture}(c).
Previous self-attentions~\cite{vaswani2017attention,liu2021swin,wu2021cvt,vaswani2021scaling,zhang2020resnest,chen2020generative} in vision define tokens with pixels or patches, and gather the information along spatial dimensions.
Instead of performing attention on pixel-level or patch-level, we apply attention mechanisms on the transpose of patch-level tokens. 
To obtain global information in the spatial dimension, we set the number of heads equal to $1$.
We argue that each transposed token abstracts the global information.
In this way, channel tokens interact with global information on the channel dimension in linear spatial-wise complexity, as shown in Figure~\ref{fig:teaser}(b).

Simply transposing the feature can obtain a vanilla channel-level attention with a complexity of $O(6PC^2)$.
To further reduce the computational complexity, we group channels into multiple groups and perform self-attention within each group.
Formally, let $N_g$ denotes the number of groups and $C_g$ denotes the number of channels in each group, we have $C = N_g * C_g$.
In this way, our channel group attention is global, with image-level tokens interacting across a group of channels.
It is defined as:
\begin{align}
\mathcal{A}_{channel}(\mathbf{Q}, \mathbf{K}, \mathbf{V}) & = 
 \{\mathcal{A}_{group}(\mathbf{Q}_i, \mathbf{K}_i, \mathbf{V}_i)^T\}_{i=0}^{N_g} \notag \\
\mathcal{A}_{group}(\mathbf{Q}_i, \mathbf{K}_i, \mathbf{V}_i) & = \mathrm{softmax} \left[\frac{\mathbf{Q}_i^\mathrm{T}\mathbf{K}_i}{\sqrt{C_g}}\right]\mathbf{V}_i^T
\label{eq:channel-attention}
\end{align}
where $\mathbf{Q}_i, \mathbf{K}_i, \mathbf{V}_i \in \mathbb{R}^{P \times C_g}$ are grouped channel-wise image-level queries, keys, and values.
Note that although we transpose the tokens in channel attention, the projection layers $\mathbf{W}$ and the scaling factor $\frac{1}{\sqrt{C_g}}$ remain performed and computed along the channel dimension, rather than the spatial one.
Considering that the number of spatial patches varies with the image size, the above design ensures our model can generalize to any image size.

\noindent \textbf{Complexity analysis.}~~
Our channel group attention is performed on the image-level tokens across the channel dimension. 
Compared to window attention that produces an attention map with size $P_w \times P_w$, the channel-wise attention map is of $C_g \times C_g$-dimensional.
The overall computational complexity of our model includes $O(2PC(P_w + C_g))$ for window and channel attentions, $O(8PC^2)$ for linear projections, and $O(16PC^2)$ for FFNs (expand ratio is 4). It can be seen that our dual attention is computationally efficient with linear complexity to both the spatial and channel dimensions.
FFN dominates the number of FLOPs and model parameters. Considering our dual attention has both channel-wise and spatial-wise interactions, in this work, we conduct an initial exploration to show the potential of the pure-attention structure without FFNs. Details can be found in Appendix.

\noindent \textbf{Global interactions in channel attention.}~~
Channel attention naturally captures global information and interactions for visual recognition tasks.
(i) After transposing the feature, each channel token itself is global on the spatial dimension, providing a global view of the image.
(ii) Given $C_g$ tokens with dimension $P$, the $C_g \times C_g$-dimensional attention map is computed by involving all spatial locations, \ie, $(C_g\times P) \cdot (P\times C_g)$.
(iii) With such a global attentive map, channel attention fuses multiple global views of the image dynamically, producing new global tokens and passing the information to the following spatial-wise layers.
Compared to spatial-wise global attentions~\cite{touvron2021training,dosovitskiy2020image} that perform interactions across spatial locations, the information exchange of our channel self-attention is imposed from a global perspective rather than a patch-wise one, complementing the spatial window attention. Detailed analysis can be found in Sec.~\ref{sec:global_analysis}.

\subsection{Model Instantiation}
\label{sec:model_config}
In this work, we follow the design strategy suggested by previous works~\cite{liu2021swin,yang2021focal}.
Take an image with $H \times W$, a $C$-dimensional feature with a resolution of $\frac{H}{4} \times \frac{W}{4}$ is obtained after the first patch embedding layer. And its resolution is further reduced into $\frac{H}{8} \times \frac{W}{8}$, $\frac{H}{16} \times \frac{W}{16}$, and $\frac{H}{32} \times \frac{W}{32}$ with the feature dimension increasing to $2C$, $4C$, and $8C$ after the other three patch embedding layer, respectively.
Here, our patch embedding layer is implemented by stride convolution. The convolutional kernels and stride values of our four patch embedding layers are $\{7, 2, 2, 2\}$ and $\{4, 2, 2, 2\}$, respectively. 

We consider three different network configurations for image classification, objection detection, and segmentation:
\begin{itemize}
\footnotesize
\vspace{-4pt}
    \item \modelInTable-Tiny: $C=96, L=\{1,1,3,1\}, N_g=N_h=\{3,6,12,24\}$
    \item \modelInTable-Small: $C=96, L=\{1,1,9,1\}, N_g=N_h=\{3,6,12,24\}$
    \item \modelInTable-Base: $C=128, L=\{1,1,9,1\}, N_g=N_h=\{4,8,16,32\}$,
\vspace{-4pt}
\end{itemize} where $L$ is the layer numbers, $N_g$ is the number of groups in channel attention, and $N_h$ is the number of heads in window attention for each stage.

When more training data involved, we further scale up \model to large, huge, and giant size to validate the scaling ability of the proposed architecture for image classification:
\begin{itemize}
\footnotesize
\vspace{-4pt}
    \item \modelInTable-Large: $C=192, L=\{1,1,9,1\}, N_g=N_h=\{6,12,24,48\}$
    \item \modelInTable-Huge: $C=256, L=\{1,1,9,1\}, N_g=N_h=\{8,16,32,64\}$
    \item \modelInTable-Giant: $C=384, L=\{1,1,12,3\}, N_g=N_h=\{12,24,48,96\}$.
\vspace{-4pt}
\end{itemize}
See Appendix for more details of model configurations.

\begin{figure}[t]
    \centering
    \begin{tabular}{cc|ccccccc}
            \centering
                \includegraphics[width=.1\columnwidth]{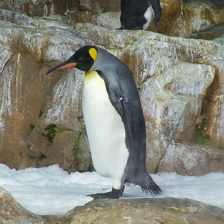} &
                \includegraphics[width=.1\columnwidth]{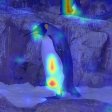}~ &
                ~\includegraphics[width=.1\columnwidth]{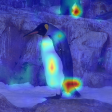} &
                \includegraphics[width=.1\columnwidth]{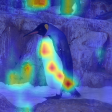} &
                \includegraphics[width=.1\columnwidth]{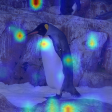} &
                \includegraphics[width=.1\columnwidth]{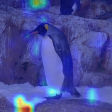} &
                \includegraphics[width=.1\columnwidth]{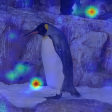} &
                \includegraphics[width=.1\columnwidth]{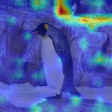} &
                \includegraphics[width=.1\columnwidth]{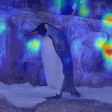} \\
                \includegraphics[width=.1\columnwidth]{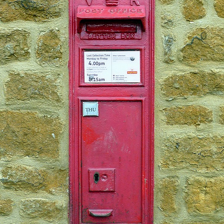} &
                \includegraphics[width=.1\columnwidth]{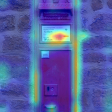}~ &
                ~\includegraphics[width=.1\columnwidth]{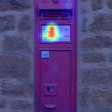} &
                \includegraphics[width=.1\columnwidth]{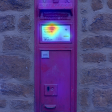} &
                \includegraphics[width=.1\columnwidth]{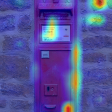} &
                \includegraphics[width=.1\columnwidth]{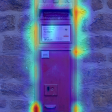} &
                \includegraphics[width=.1\columnwidth]{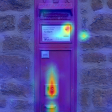} &
                \includegraphics[width=.1\columnwidth]{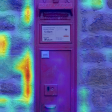} &
                \includegraphics[width=.1\columnwidth]{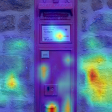} \\
                \includegraphics[width=.1\columnwidth]{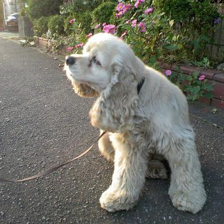} &
                \includegraphics[width=.1\columnwidth]{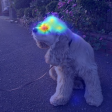}~ &
                ~\includegraphics[width=.1\columnwidth]{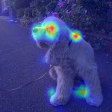} &
                \includegraphics[width=.1\columnwidth]{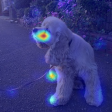} &
                \includegraphics[width=.1\columnwidth]{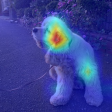} &
                \includegraphics[width=.1\columnwidth]{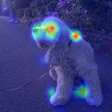} &
                \includegraphics[width=.1\columnwidth]{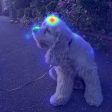} &
                \includegraphics[width=.1\columnwidth]{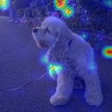} &
                \includegraphics[width=.1\columnwidth]{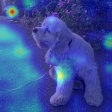} \\
                \tabincell{c}{\scriptsize Original \\ \scriptsize image} & \tabincell{c}{\scriptsize Output \\ \scriptsize featuremap} & \multicolumn{7}{c}{
                \tabincell{c}{\scriptsize Input featuremaps before the channel attention \\ \scriptsize (top-7 channels attending to the output channel, sorted by attention score)}
                }
    \end{tabular}
    \vspace{-8pt}
    \caption{Illustrating how channel attention gathering global information by visualizing attended feature maps.
The first and second columns denote the original image and a feature channel after our channel attention; while the other columns are the channel tokens with top-7 highest attention scores before the attention. Channel attention is able to select globally important regions and suppress background regions for better recognition. The third network stage of the classification model is used for visualization.}
    \vspace{-8pt}
    \label{fig:channel_vis}
\end{figure}

\begin{figure}[t]
\centering
\footnotesize
~~~~DaViT~~~~~Swin~~~~~DeiT~~~~~DaViT~~~~~Swin~~~~~DeiT~~~~~DaViT~~~~~Swin~~~~~DeiT~~
\\
\vspace{0.5mm}
\rotatebox{90}{~Stage 1}
\includegraphics[width=.1\columnwidth]{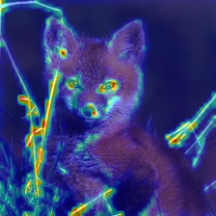}\hspace{-0.2pt}
\includegraphics[width=.1\columnwidth]{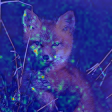}\hspace{-0.2pt}
\includegraphics[width=.1\columnwidth]{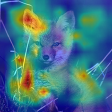}\hspace{1.5pt}
\includegraphics[width=.1\columnwidth]{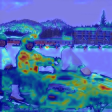}\hspace{-0.2pt}
\includegraphics[width=.1\columnwidth]{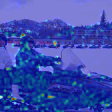}\hspace{-0.2pt}
\includegraphics[width=.1\columnwidth]{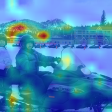}\hspace{1.5pt}
\includegraphics[width=.1\columnwidth]{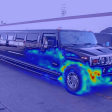}\hspace{-0.2pt}
\includegraphics[width=.1\columnwidth]{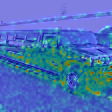}\hspace{-0.2pt}
\includegraphics[width=.1\columnwidth]{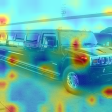}
\\
\vspace{0.5mm}
\rotatebox{90}{~Stage 2}
\includegraphics[width=.1\columnwidth]{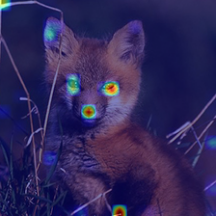}\hspace{-0.2pt}
\includegraphics[width=.1\columnwidth]{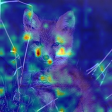}\hspace{-0.2pt}
\includegraphics[width=.1\columnwidth]{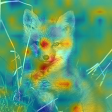}\hspace{1.5pt}
\includegraphics[width=.1\columnwidth]{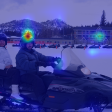}\hspace{-0.2pt}
\includegraphics[width=.1\columnwidth]{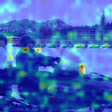}\hspace{-0.2pt}
\includegraphics[width=.1\columnwidth]{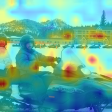}\hspace{1.5pt}
\includegraphics[width=.1\columnwidth]{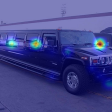}\hspace{-0.2pt}
\includegraphics[width=.1\columnwidth]{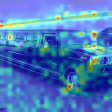}\hspace{-0.2pt}
\includegraphics[width=.1\columnwidth]{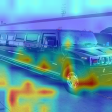}
\\
\vspace{0.5mm}
\rotatebox{90}{~Stage 3}
\includegraphics[width=.1\columnwidth]{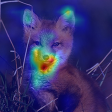}\hspace{-0.2pt}
\includegraphics[width=.1\columnwidth]{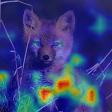}\hspace{-0.2pt}
\includegraphics[width=.1\columnwidth]{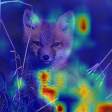}\hspace{1.5pt}
\includegraphics[width=.1\columnwidth]{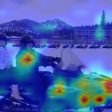}\hspace{-0.2pt}
\includegraphics[width=.1\columnwidth]{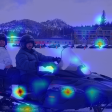}\hspace{-0.2pt}
\includegraphics[width=.1\columnwidth]{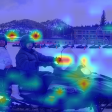}\hspace{1.5pt}
\includegraphics[width=.1\columnwidth]{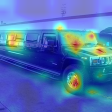}\hspace{-0.2pt}
\includegraphics[width=.1\columnwidth]{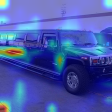}\hspace{-0.2pt}
\includegraphics[width=.1\columnwidth]{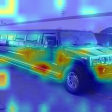}
\\
\vspace{0.5mm}
\rotatebox{90}{~Stage 4}
\includegraphics[width=.1\columnwidth]{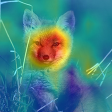}\hspace{-0.2pt}
\includegraphics[width=.1\columnwidth]{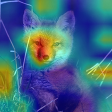}\hspace{-0.2pt}
\includegraphics[width=.1\columnwidth]{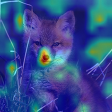}\hspace{1.5pt}
\includegraphics[width=.1\columnwidth]{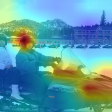}\hspace{-0.2pt}
\includegraphics[width=.1\columnwidth]{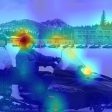}\hspace{-0.2pt}
\includegraphics[width=.1\columnwidth]{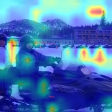}\hspace{1.5pt}
\includegraphics[width=.1\columnwidth]{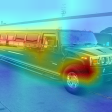}\hspace{-0.2pt}
\includegraphics[width=.1\columnwidth]{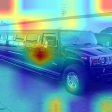}\hspace{-0.2pt}
\includegraphics[width=.1\columnwidth]{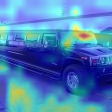}
\\
\vspace{-4pt}
\caption{Feature map visualization of our DaViT, Swin~\cite{liu2021swin}, and DeiT~\cite{touvron2021training} at four different network stages (from top to bottom representing stages 1 to 4). 
DeiT~\cite{touvron2021training} is divided into 4 stages by ${2,2,6,2}$ layers, respectively.
For all network stages, a random feature channel is visualized.
}
\vspace{-8pt}
\label{fig:global_vis}
\end{figure}

\vspace{-4pt}
\section{Analysis}
\vspace{-4pt}
\label{sec:global_analysis}

\noindent \textbf{Interpretation of global interactions.}~~
Global interactions in transformers can be summarized into different types.
Vanilla ViT~\cite{dosovitskiy2020image} and DeiT~\cite{touvron2021training} perform information exchange between different patches among the whole image; Focal Transformer~\cite{yang2021focal} proposes interactions between tokens with different scales to get a larger receptive field; PVT~\cite{wang2021pvtv2} leverages a spatial reduction mechanism to obtain a coarse approximation of global attention; Swin~\cite{liu2021swin} stacks multiple hierarchical layers to get global information eventually.
%
Unlike them, a single block with our channel attention is able to learn the global interactions from another perspective by taking all spatial positions into account when computing the attention scores, as in $(C_g\times P) \cdot (P\times C_g)$.
It captures the information from multiple global tokens, which represent different abstract views of the entire image.
For example, different channels may contain information from different parts of an object; such part information can be aggregated into a global view.

Figure~\ref{fig:channel_vis} illustrates how our channel attention works by visualizing the featuremaps before and after the attention. For each image in the first column, we randomly choose an output channel (the second column) in the third stage of the network and its corresponding top-7 relevant input channels for visualization.
We can see that the channel attention fuses information from multiple tokens, selects globally important regions, and suppresses unimportant regions.

\noindent \textbf{Relation to Swin and DeiT.}~~
To show the effectiveness of our dual self-attention, we make detailed comparisons with two representative clean baselines: Swin~\cite{liu2021swin} with fine-grained local features, and DeiT~\cite{touvron2021training} with global coarse features.
Note that though our work and Swin both use window attention as a network element, the key design of Swin, \ie, the use of ``shifted window partitions'' between successive layers to increase the receptive field, is not utilized in our work.
We also simplifies its relative position encoding~\cite{chu2021conditional} with a depth-wise convolution before layernorm to get a cleaner structure for arbitrary input sizes.
To further keep our architecture clean and efficient, we do not use additional operators like ``Overlapping Patch''~\cite{wu2021cvt,wang2021pvtv2} and ``ConvFFN''~\cite{yuan2021hrformer,xie2021segformer}.
See Appendix for detailed throughput of our model compared with Swin.

Figure~\ref{fig:global_vis} shows the effectiveness of our channel group attention.
We randomly visualize a feature channel in each stage of DaViT, Swin~\cite{liu2021swin}, and DeiT~\cite{touvron2021training}. We observe that:
(i) Swin captures fine-grained details but no focus in the first two stages as it lacks global information. It can not focus on the main object until the last stage.
(ii) DeiT learns coarse-grained global features over the image but loses details hence difficult to focus on the main content.
(iii) Our DaViT captures both short-range and long-range visual dependencies by combining two types of self-attention. It shows strong global modeling capabilities by finding out fine-grained details of the main content in stage 1, and further focusing on some keypoints in stage 2.
It then gradually refines the regions of interest from both global and local perspectives for final recognition.

\begin{table}[t]
\footnotesize
\centering
\setlength{\tabcolsep}{2pt}
\renewcommand\arraystretch{1.1}
\caption{Comparison of image classification on ImageNet-1K for different models. 
All models are trained and evaluated with $224 \times 224$ resolution on ImageNet-1K by default, unless otherwise noted. 
For fair comparison, token labeling~\cite{ZihangJiang2021AllTM} and distillation~\cite{touvron2021training} are not used for all models and their counterparts.
$\dag$ and $\ddag$ denote the model is evaluated with resolution of  $384 \times 384$ and $512 \times 512$, respectively.
}
\vspace{-4pt}
\resizebox{0.49\linewidth}{!}{
    \begin{tabular}{l|ccc}
    \shline
    \multirow{2}{*}{Model} & \#Params & FLOPs & Top-1 \\
    & (M) & (G) & (\%) \\
    \shline
    ResNet-50~\cite{he2016deep} & 25.0 & 4.1 & 76.2 \\
    DeiT-Small/16~\cite{touvron2021training} & 22.1 & 4.5 & 79.8 \\
    PVT-Small~\cite{wang2021pyramid}  & 24.5 & 3.8 & 79.8 \\
    ConvMixer-768/32~\cite{anonymous2022patches} & 21.1 & -- & 80.2 \\
    CrossViT-Small~\cite{chen2021crossvit} & 26.7 & 5.6 & 81.0 \\
    Swin-Tiny~\cite{liu2021swin}   & 28.3 & 4.5 & 81.2 \\
    CvT-13~\cite{wu2021cvt}        & 20.0 & 4.5 & 81.6 \\
    CoAtNet-0~\cite{dai2021coatnet} & 25.0 & 4.2 & 81.6 \\
    CaiT-XS-24~\cite{touvron2021going} & 26.6 & 5.4 & 81.8 \\
    ViL-Small~\cite{zhang2021multi}   & 24.6 & 5.1 & 82.0 \\
    PVTv2-B2~\cite{wang2021pvtv2} & 25.4 & 4.0 & 82.0 \\
    UFO-ViT-S~\cite{song2021ufo} & 21.0 & 3.7 & 82.0 \\
    Focal-Tiny~\cite{yang2021focal}   & 29.1 & 4.9 & 82.2 \\
    \rowcolor{Gray}
    \modelInTable-Tiny~(Ours)         & 28.3 & 4.5 & \textbf{82.8} \\
    \hline
    ResNet-101~\cite{he2016deep}  & 45.0 & 7.9 & 77.4 \\
    PVT-Medium~\cite{wang2021pyramid} & 44.2 & 6.7 & 81.2 \\
    CvT-21~\cite{wu2021cvt}  & 32.0 & 7.1 & 82.5 \\
    UFO-ViT-M~\cite{song2021ufo}  & 37.0 & 7.0 & 82.8 \\
    Swin-Small~\cite{liu2021swin}    &  49.6 & 8.7 & 83.1 \\
    ViL-Medium~\cite{zhang2021multi}  & 39.7 & 9.1 & 83.3 \\
    CaiT-S36~\cite{touvron2021going} & 68.0 & 13.9 & 83.3 \\
    CoAtNet-1~\cite{dai2021coatnet} & 42.0 & 8.4 & 83.3 \\
    Focal-Small~\cite{yang2021focal} & 51.1 & 9.1 & 83.5 \\
    CSwin-S~\cite{dong2021cswin} & 35.0 & 6.9 & 83.6 \\
    VAN-Large~\cite{guo2022visual} & 44.8 & 9.0 & 83.9 \\
    UniFormer-B~\cite{li2022uniformer} & 50.0 & 8.3 & 83.9 \\
    \rowcolor{Gray}
    \modelInTable-Small~(Ours) & 49.7 & 8.8 & \textbf{84.2} \\
    \shline
    \end{tabular}
    }
    \hfill
\resizebox{0.49\linewidth}{!}{
    \begin{tabular}{l|ccc}
    \shline
    \multirow{2}{*}{Model} & \#Params & FLOPs & Top-1 \\
    & (M) & (G) & (\%) \\
    \shline
    ResNet-152~\cite{he2016deep}  & 60.0 & 11.0 & 78.3 \\
    PVT-Large~\cite{wang2021pyramid} & 61.4 & 9.8 & 81.7 \\
    DeiT-Base/16~\cite{touvron2021training}& 86.7 & 17.4 & 81.8 \\
    CrossViT-Base~\cite{chen2021crossvit} & 104.7 & 21.2 & 82.2 \\
    T2T-ViT-24~\cite{yuan2021tokens}  & 64.1 & 14.1 & 82.3 \\
    CPVT-Base~\cite{chu2021conditional}  & 88.0 & 17.6 & 82.3 \\
    TNT-Base~\cite{han2021transformer} & 65.6 & 14.1 & 82.8\\
    ViL-Base~\cite{zhang2021multi} & 55.7 & 13.4 & 83.2 \\
    UFO-ViT-B~\cite{song2021ufo} & 64.0 & 11.9 & 83.3 \\
    Swin-Base~\cite{liu2021swin}   & 87.8 & 15.4 & 83.4 \\
    CaiT-M24~\cite{touvron2021going} & 185.9 & 36.0 & 83.4 \\
    NFNet-F0~\cite{AndrewBrock2021HighPerformanceLI}  & 71.5 & 12.4 & 83.6 \\
    PVTv2-B5~\cite{wang2021pvtv2} & 82.0 & 11.8 & 83.8\\
    Focal-Base~\cite{yang2021focal}  & 89.8 & 16.0 & 83.8 \\
    CoAtNet-2~\cite{dai2021coatnet} & 75.0 & 15.7 & 84.1 \\
    CSwin-B~\cite{dong2021cswin} & 78.0 & 15.0 & 84.2 \\
    \rowcolor{Gray}
    \modelInTable-Base~(Ours)~~~~\hspace{1.5pt} & 87.9 & 15.5 & \textbf{84.6} \\
    \hline
    \multicolumn{4}{c}{\small Pre-trained on ImageNet-22k} \\
    \hline
    Swin-Large~\cite{liu2021swin} $\dag $ & 197.0 & 103.9 & 86.4 \\
    CSWin-B~\cite{dong2021cswin} $\dag$ & 78.0 & 47.0 & 87.0 \\
    CSWin-L~\cite{dong2021cswin} $\dag$ & 173.0 & 96.8 & 87.5 \\
    CoAtNet-3~\cite{dai2021coatnet} $\dag$ & 168.0 & 107.4 & 87.6 \\
    \rowcolor{Gray}
    \modelInTable-Base~(Ours) $\dag$ & 87.9 & 46.4 & 86.9 \\
    \rowcolor{Gray}
    \modelInTable-Large~(Ours) $\dag$ & 196.8 & 103.0 & 87.5 \\
    \hline
    \multicolumn{4}{c}{\small Pre-trained on 1.5B image and text pairs} \\
    \hline
    \rowcolor{Gray}
    \modelInTable-Huge~(Ours) $\ddag$ & 362& 334& 90.2 \\
    \rowcolor{Gray}
    \modelInTable-Giant~(Ours) $\ddag$ & 1437&1038 & \textbf{90.4} \\
    \hline
    \shline
    \end{tabular}
    }
    \label{tab:image_classification}
    \vspace{-8pt}
\end{table}

\noindent \textbf{Relation to channel-wise attention in CNNs.}~~
Traditional channel-wise attentions blocks, \eg, SENet~\cite{hu2018squeeze} and ECANet~\cite{wang2020eca}, adaptively reweight the feature maps by global averaged features among channels. In contrast, our channel group self-attention performs dynamic feature fusion across global tokens (different global views of the entire image) in transformers.
In this way, our channel group self-attention is more powerful. We make quantitative comparisons by replacing the channel self-attention in our tiny model with SE~\cite{hu2018squeeze} and ECA~\cite{wang2020eca} blocks. See Appendix for detailed results.

\section{Experiments}
We conduct experiments on ImageNet-1K image classification~\cite{deng2009imagenet}, COCO object detection~\cite{lin2014microsoft}, and ADE20K semantic segmentation~\cite{zhou2017scene}. Neither token labeling~\cite{ZihangJiang2021AllTM} nor distillation~\cite{touvron2021training} is used in all experiments and comparisons.

\begin{table}[t!]
\footnotesize
\centering
\caption{Comparisons with CNN and Transformer baselines and SoTA methods on COCO object detection. The box mAP ($AP^b$) and mask mAP ($AP^m$) are reported for RetinaNet and Mask R-CNN trained with $1\times$ schedule. FLOPs are measured by $800 \times 1280$. More detailed comparisons with $3\times$ schedule are in Table~\ref{tab:object_detection_3x}.}
\vspace{-4pt}
\setlength{\tabcolsep}{10pt}
\renewcommand\arraystretch{1.1}
\resizebox{0.7\linewidth}{!}{
    \begin{tabular}{l|c|c|cc}
    \shline
     \multirow{2}{*}{Backbone} & FLOPs & RetinaNet & \multicolumn{2}{c}{Mask R-CNN} \\
     & (G) & $AP^{b}$ & $AP^{b}$ & $AP^{m}$\\
    \shline
    ResNet-50~\cite{he2016deep} & 239/260 & 36.3 & 38.0 & 34.4 \\
    PVT-Small~\cite{wang2021pyramid} &226/245& 40.4 & 40.4 & 37.8 \\
    ViL-Small~\cite{zhang2021multi} &252/174& 41.6 & 41.8 & 38.5 \\
    Swin-Tiny~\cite{liu2021swin} &245/264& 42.0 & 43.7 & 39.8 \\
    Focal-Tiny~\cite{yang2021focal}     &265/291& 43.7 & 44.8 & 41.0 \\
    \rowcolor{Gray}
    \modelInTable-Tiny~(Ours) &244/263& \textbf{44.0} & \textbf{45.0} & \textbf{41.1} \\
    \hline
    ResNeXt101-32x4d~\cite{xie2017aggregated} &319/340& 39.9 & 41.9 & 37.5 \\
    PVT-Medium~\cite{wang2021pyramid} &283/302& 41.9 & 42.0 & 39.0 \\
    ViL-Medium~\cite{zhang2021multi} &339/261& 42.9 & 43.4 & 39.7 \\
    Swin-Small~\cite{liu2021swin}  &335/354& 45.0 & 46.5 & 42.1 \\  
    Focal-Small~\cite{yang2021focal} &367/401& 45.6 & 47.4 & 42.8 \\
    \rowcolor{Gray}
    \modelInTable-Small~(Ours) &332/351& \textbf{46.0} & \textbf{47.7} & \textbf{42.9} \\
    \hline
    ResNeXt101-64x4d~\cite{xie2017aggregated} &473/493& 41.0 & 42.8 & 38.4 \\    
    PVT-Large~\cite{wang2021pyramid}  &345/364& 42.6 & 42.9 & 39.5 \\
    ViL-Base~\cite{zhang2021multi} &443/365& 44.3 & 45.1 & 41.0 \\
    Swin-Base~\cite{liu2021swin} &477/496& 45.0 & 46.9 & 42.3 \\  
    Focal-Base~\cite{yang2021focal} &514/533& 46.3 & 47.8 & 43.2 \\
    \rowcolor{Gray}
    \modelInTable-Base~(Ours) &471/491& \textbf{46.7} & \textbf{48.2} & \textbf{43.3} \\
    \shline
    \end{tabular}}
    \label{tab:object_detection}
    \vspace{-8pt}
\end{table}

\begin{table*}[t]
\begin{center}
\caption{COCO object detection and segmentation results with RetinaNet~\cite{lin2017focal} and Mask R-CNN~\cite{he2016deep}. All models are trained with $3\times$ schedule and multi-scale inputs. The numbers before and after ``/'' at column 2 and 3 are the model size and complexity for RetinaNet and Mask R-CNN, respectively. FLOPs are measured by $800 \times 1280$.}
\vspace{-4pt}
\resizebox{\linewidth}{!}{
\setlength{\tabcolsep}{2.1pt}
\renewcommand\arraystretch{1.1}
\footnotesize
\begin{tabular}{l|cc|cccccc|cccccc}
\shline
\multirow{2}{*}{Backbone} & \#Params & FLOPs &  \multicolumn{6}{c}{RetinaNet 3x} & \multicolumn{6}{c}{Mask R-CNN 3x}\\
 & (M) & (G) & $AP^b$ & $AP^b_{50}$ & $AP^b_{75}$ & $AP_{S}$ & $AP_{M}$ & $AP_{L}$ & $AP^b$ & $AP^b_{50}$ & $AP^b_{75}$ & $AP^m$ & $AP^m_{50}$ & $AP^m_{75}$\\
\shline
ResNet50~\cite{he2016deep} & 37.7/44.2 & 239/260 & 39.0 & 58.4 & 41.8 & 22.4 & 42.8 & 51.6 & 41.0 & 61.7 & 44.9 & 37.1 & 58.4 & 40.1 \\
PVT-Small\cite{wang2021pyramid} & 34.2/44.1 & 226/245 & 42.2 & 62.7 & 45.0 & 26.2 & 45.2 & 57.2 & 43.0 & 65.3 & 46.9 & 39.9 & 62.5 & 42.8 \\
ViL-Small~\cite{zhang2021multi} & 35.7/45.0 & 252/174 & 42.9 & 63.8 & 45.6 & 27.8 & 46.4 & 56.3 & 43.4 & 64.9 & 47.0 & 39.6 & 62.1 & 42.4 \\
Swin-Tiny~\cite{liu2021swin} & 38.5/47.8 & 245/264 & 45.0 & 65.9 & 48.4 & 29.7 & 48.9 & 58.1 & 46.0 & 68.1 & 50.3 & 41.6 & 65.1 & 44.9 \\
Focal-Tiny~\cite{yang2021focal} & 39.4/48.8 & 265/291 & 45.5 & 66.3 & 48.8 & 31.2 & {49.2} & {58.7} & {47.2} & {69.4} & {51.9} & {42.7} & {66.5} & {45.9} \\

\rowcolor{Gray}
\modelInTable-Tiny (Ours) & 38.5/47.8 & 244/263 & \textbf{46.5} & \textbf{68.1} & \textbf{49.6} & \textbf{32.3} & \textbf{50.6} & \textbf{59.9} & \textbf{47.4} & \textbf{69.5} & \textbf{52.0} & \textbf{42.9} & \textbf{66.8} & \textbf{46.4} \\

\hline
ResNeXt101-32x4d~\cite{xie2017aggregated} & 56.4/62.8 & 319/340 & 41.4 & 61.0 & 44.3 & 23.9 & 45.5 & 53.7 & 44.0 & 64.4 & 48.0 & 39.2 & 61.4 & 41.9 \\
PVT-Medium~\cite{wang2021pyramid} & 53.9/63.9 & 283/302 & 43.2 & 63.8 & 46.1 & 27.3 & 46.3 & 58.9 & 44.2 & 66.0 & 48.2 & 40.5 & 63.1 & 43.5 \\
ViL-Medium~\cite{zhang2021multi} & 50.8/60.1 & 339/261 & 43.7 & 64.6 & 46.4 & 27.9 & 47.1 & 56.9 & 44.6 & 66.3 & 48.5 & 40.7 & 63.8 & 43.7 \\
Swin-Small~\cite{liu2021swin} & 59.8/69.1 & 335/354 & 46.4 & 67.0 & 50.1 & 31.0 & 50.1 & 60.3  & 48.5 & 70.2 & 53.5 & 43.3 & 67.3 & 46.6 \\
Focal-Small~\cite{yang2021focal}  & 61.7/71.2 & 367/401 & {47.3} & {67.8} & {51.0} & {31.6} & {50.9} & {61.1} & {48.8} & {70.5} & {53.6} & {43.8} & {67.7} & {47.2} \\

\rowcolor{Gray}
\modelInTable-Small (Ours) & 59.9/69.2 & 332/351 & \textbf{48.2} & \textbf{69.7} & \textbf{51.7} & \textbf{32.7} & \textbf{52.2} & \textbf{62.9} & \textbf{49.5} & \textbf{71.4} & \textbf{54.7} & \textbf{44.3} & \textbf{68.4} & \textbf{47.6} \\
\hline
ResNeXt101-64x4d~\cite{xie2017aggregated} & 95.5/102 & 473/493 & 41.8 & 61.5 & 44.4 & 25.2 & 45.4 & 54.6 & 44.4 & 64.9 & 48.8 & 39.7 & 61.9 & 42.6 \\
PVT-Large\cite{wang2021pyramid} & 71.1/81.0 & 345/364 & 43.4 & 63.6 & 46.1 & 26.1 & 46.0 & 59.5 & 44.5 & 66.0 & 48.3 & 40.7 & 63.4 & 43.7 \\
ViL-Base~\cite{zhang2021multi} & 66.7/76.1 & 443/365 & 44.7 & 65.5 & 47.6 & 29.9 & 48.0 & 58.1 & 45.7 & 67.2 & 49.9 & 41.3 & 64.4 & 44.5 \\
Swin-Base~\cite{liu2021swin} & 98.4/107.0 & 477/496 & 45.8 & 66.4 & 49.1 & 29.9 & 49.4 & 60.3 & 48.5 & 69.8 & 53.2 & 43.4 & 66.8 & 46.9 \\
Focal-Base~\cite{yang2021focal} & 100.8/110.0 & 514/533 & {46.9} & {67.8} & {50.3} & {31.9} & {50.3} & {61.5} & {49.0} & {70.1} & {53.6} & {43.7} & {67.6} & {47.0} \\
\rowcolor{Gray}
\modelInTable-Base (Ours) & 98.5/107.3 & 471/491 & \textbf{48.7} & \textbf{70.0} & \textbf{52.3} & \textbf{33.7} & \textbf{52.8} & \textbf{62.9} & \textbf{49.9} & \textbf{71.5} & \textbf{54.6} & \textbf{44.6} & \textbf{68.8} & \textbf{47.8} \\
\shline
\end{tabular}}
\end{center}
\label{tab:object_detection_3x}
\vspace{-8pt}
\end{table*}

\begin{table}
\setlength{\tabcolsep}{8pt}
\renewcommand\arraystretch{1.1}
\centering
\footnotesize
\caption{Comparison with SoTA methods for semantic segmentation on ADE20K~\cite{zhou2017scene} val set.
Single-scale evaluation is used. FLOPs are measured by $512 \times 2048$.
}
\vspace{-4pt}
\resizebox{0.7\linewidth}{!}{
  \begin{tabular}{ll|ccc}
    \shline
    \multirow{2}{*}{Backbone} & \multirow{2}{*}{Method} & \#Params & FLOPs & mIoU \\
    && (M) & (G) & (\%)\\
    \shline	 
    Swin-Tiny~\cite{liu2021swin} & UperNet~\cite{xiao2018unified} & 60 & 945 & 44.5 \\
    PVT-Large~\cite{wang2021pyramid} & SemanticFPN~\cite{lin2017feature} & 65 & 318 & 44.8 \\
    HRNet-w48~\cite{wang2020deep} & OCRNet~\cite{yuan2019object} & 71 & 664 & 45.7 \\
    Focal-Tiny~\cite{yang2021focal} & UperNet~\cite{xiao2018unified} & 62 & 998 & 45.8 \\
    XCiT-S12/16~\cite{ali2021xcit} & UperNet~\cite{xiao2018unified} & 52 & -- & 45.9 \\
    Twins-SVT-Small~\cite{chu2021twins} & UperNet~\cite{xiao2018unified} & 54 & 912 & 46.2 \\
     \rowcolor{Gray} 
    \modelInTable-Tiny~(Ours) & UperNet~\cite{xiao2018unified} & 60 & 940 & \textbf{46.3}  \\
    \hline
    ResNet-101~\cite{he2016deep} & UperNet~\cite{xiao2018unified} & 86 & 1029 & 44.9 \\
    XCiT-S24/16~\cite{ali2021xcit} & UperNet~\cite{xiao2018unified} & 73 & -- & 46.9 \\
    Swin-Small~\cite{liu2021swin} & UperNet~\cite{xiao2018unified} & 81 & 1038 & 47.6 \\
    Twins-SVT-Base~\cite{chu2021twins} & UperNet~\cite{xiao2018unified} & 88 & 1044 & 47.7 \\
    Focal-Small~\cite{yang2021focal} & UperNet~\cite{xiao2018unified} & 85 & 1130 & 48.0 \\
    ResNeSt-200~\cite{zhang2020resnest} & DLab.v3+~\cite{chen2018encoder} & 88 & 1381 & 48.4 \\
     \rowcolor{Gray}
    \modelInTable-Small~(Ours) & UperNet~\cite{xiao2018unified} & 81 & 1030 & \textbf{48.8} \\
    \hline
    Swin-Base~\cite{liu2021swin} & UperNet~\cite{xiao2018unified} & 121 & 1188 & 48.1 \\
    XCiT-M24/8~\cite{ali2021xcit} & UperNet~\cite{xiao2018unified} & 109 & -- & 48.4 \\
    Twins-SVT-Large~\cite{chu2021twins} & UperNet~\cite{xiao2018unified} & 133 & 1188 & 48.8 \\
    ViT-Hybrid~\cite{ranftl2021vision} & DPT~\cite{ranftl2021vision} & 124 & 1231 & 49.0 \\
    Focal-Base~\cite{yang2021focal} & UperNet~\cite{xiao2018unified} & 126 & 1354 & 49.0 \\
     \rowcolor{Gray} 
    \modelInTable-Base~(Ours) & UperNet~\cite{xiao2018unified} & 121 & 1175 & \textbf{49.4} \\
    \shline
  \end{tabular} 
  }
  \label{tab:semantic_segmentation}
\end{table}

\subsection{Image Classification}

We compare different methods on ImageNet-1K~\cite{deng2009imagenet}. We implement our \model on the timm framework~\cite{wightman2019pytorch}.
Following~\cite{liu2021swin,lin2017focal,wu2021cvt,zhang2021multi}, we use the same set of data augmentation and regularization strategies used in~\cite{touvron2021training} after excluding repeated augmentation~\cite{berman2019multigrain,hoffer2020augment} and exponential moving average (EMA)~\cite{polyak1992acceleration}. 
We train all the models for $300$ epochs with a batch size 2048 and use AdamW~\cite{loshchilov2017decoupled} as the optimizer.
The weight decay is set to $0.05$ and the maximal gradient norm is clipped to $1.0$.
We use a simple triangular learning rate schedule~\cite{smith2019super} as in ~\cite{anonymous2022patches}.
The stochastic depth drop rates are set to $0.1$, $0.2$, and $0.4$ for our tiny, small, and base models, respectively. During training, we crop images randomly to $224 \times 224$, while a center crop is used during evaluation on the validation set.
%

In Table~\ref{tab:image_classification}, we summarize the results for baseline models and current state-of-the-art models on the image classification task. We can find our \model achieves new state-of-the-art and consistently outperforms other methods with similar model size (\#Params.) and computational complexity (GFLOPs). 
Specifically, \modelInTable-Tiny, Small, and Base improve over the Swin Transformer~\cite{liu2021swin} by 1.5\%, 1.1\%, and 1.2\%, respectively.
%
%
Notably, our \modelInTable-Small with 49.7M parameters reaches 84.2\%, which surpasses all counterpart -Base models using much fewer parameters. For example, our \modelInTable-Small achieves 0.4\% and 0.8\% higher accuracy than Focal-Base and Swin-Base, respectively, using near half computations. 

Following~\cite{wu2021cvt,dosovitskiy2020image,liu2021swin}, when 13M images from ImageNet-22k~\cite{deng2009imagenet} involved for pre-training, \modelInTable-Base and \modelInTable-Large obtained 86.9\% and 87.5\% top-1 accuracy, respectively. Furthermore, when we further scale up \model with 1.5B privately collected weakly supervised image-text pairs data and pre-train \model with unified contrastive learning~\cite{yuan2021florence} approach, \modelInTable-Huge and \modelInTable-Gaint reach 90.2\% and 90.4\% top-1 accuracy on ImageNet with 362M and 1.4B parameters, respectively.

\subsection{Object Detection and Instance Segmentation}

We benchmark our models on object detection with COCO~2017~\cite{lin2014microsoft}. The pre-trained models are used as visual backbones and then plugged into two representative pipelines, RetinaNet~\cite{lin2017focal} and Mask R-CNN~\cite{he2017mask}. All models are trained on the 118k training images and results reported on the 5K validation set. We follow the standard to use two training schedules, $1\times$ schedule with $12$ epochs and $3\times$ schedule with 36 epochs. The same multi-scale training strategy as in \cite{liu2021swin} by randomly resizing the shorter side of the image to the range of $[480,800]$ is used. During training, we use AdamW~\cite{loshchilov2017decoupled} for optimization with initial learning rate $10^{-4}$ and weight decay $0.05$. We use $0.1$, $0.2$, and $0.3$ stochastic depth drop rates to regularize the training for our tiny, small, and base models, respectively. The numbers of counterparts~\cite{liu2021swin,zhang2021multi} are borrowed from~\cite{yang2021focal}.

In Table~\ref{tab:object_detection} and Table~\ref{tab:object_detection_3x}, we show the performance of our models against several state-of-the-art counterparts. The bbox mAP ($AP^b$) and mask mAP ($AP^m$) are reported.
Results of $1\times$ schedule shown in Table~\ref{tab:object_detection} have demonstrated the effectiveness of our method.
We observe substantial gains across all settings and metrics compared with several strong transformer baselines.

To have more comprehensive comparisons, we further train our models with $3\times$ schedule and show the detailed numbers, \#parameters, and associated computational costs for RetinaNet and Mask R-CNN in Table~\ref{tab:object_detection_3x}. As we can see, even for the $3\times$ schedule, our models can still achieve 1.5-2.9\% gains on RetinaNet $3\times$ and 1.0-1.4\% gains on Mask R-CNN $3\times$ over Swin Transformer models.

Moreover, from Table~\ref{tab:object_detection_3x} we observe a saturated and even degraded mAP in Swin Transformer~\cite{liu2021swin}
and Focal Transformer~\cite{yang2021focal} from small to base model, while the mAP of our model is continuously increased with larger model size, showing a better scale-up ability. Our base model outperforms the state-of-the-art~\cite{yang2021focal} by 1.8\% on RetinaNet $3\times$ and 0.9\% on Mask R-CNN $3\times$.

\subsection{Semantic Segmentation on ADE20k}
Besides the instance segmentation results above, we further evaluate our model on semantic segmentation, a task that usually requires high-resolution input and long-range interactions. We benchmark our method on ADE20K~\cite{zhou2017scene}. Specifically, we use UperNet~\cite{xiao2018unified} as the segmentation method and our \model as the backbone. We train three models with \modelInTable-Tiny, \modelInTable-Small, \modelInTable-Base, respectively. For all models, we use a standard recipe by setting the input size to $512 \times 512$ and train the model for 160k iterations with batch size 16. 
In Table~\ref{tab:semantic_segmentation}, we show the comparisons to previous works. As we can see, our tiny, small, and base models consistently outperform recent SoTAs, such as 1.2-1.8\% gains over Swin Transformers~\cite{liu2021swin} with a similar number of parameters and FLOPs.

\begin{table}[t]
\footnotesize
\centering
\caption{The effect of channel group attention at different network stages. Taking a transformer layout with all spatial window attention blocks as the baseline (n/a), we replace two spatial attention blocks at different stages with a dual attention block to show its effectiveness. The first two spatial attention blocks are selected in the third stage to compare with other stages fairly.
}
\vspace{-5pt}
\setlength{\tabcolsep}{10pt}
\renewcommand\arraystretch{1.1}
\resizebox{0.7\linewidth}{!}{
    \begin{tabular}{l|ccccc}
    \shline
    Stage~~~  & \#Params (M) & FLOPs (G) & Top-1 (\%) & Top-5 (\%)  \\
    \shline
    n/a & 28.3 & 4.6 & 81.1 & 95.6 \\
    1 & 28.3 & 4.5 & 81.7 & 95.9 \\
    2 & 28.3 & 4.5 & 82.2 & 96.1 \\
    3 & 28.3 & 4.6 & 82.1 & 96.0 \\
    4 & 28.3 & 4.6 & 81.9 & 95.9 \\
    1--4 & 28.3 & 4.6 & \textbf{82.8} & \textbf{96.2} \\
    \shline
    \end{tabular}
    }
    \label{tab:different_stages}
\end{table}

\begin{table}[t]
\footnotesize
\centering
\caption{Quantitative comparisons of different dual attention layouts on ImageNet.\hfill}
\vspace{-5pt}
\renewcommand\arraystretch{1.1}
\setlength{\tabcolsep}{8pt}
\resizebox{0.7\linewidth}{!}{
    \begin{tabular}{l|ccc}
    \shline
    Model  & \#Params (M) & FLOPs (G) & Top-1 (\%)  \\
    \shline
    Window $\rightarrow$ Channel & 28.3 & 4.5 & \textbf{82.8} \\
    Channel $\rightarrow$ Window & 28.3 & 4.5  & 82.6 \\
    Hybrid (parallel) & 28.3 & 4.5 & 82.6 \\
     \shline
    \end{tabular}
    }
    \label{tab:dual_att_arch}
    \vspace{-8pt}
\end{table}

\subsection{Ablation Study}

\noindent\textbf{Evaluation of channel group attention at different stages.}~~
We make comparisons by inserting a dual attention block at different stages of a window transformer.
From the quantitative results in Table~\ref{tab:different_stages}, we observe:
(i)  The dual attention module consistently boosts performance at each stage.
(ii) Dual attention in the second stage improves the most, as the earlier stage requires more global information. We speculate that the relatively small improvement in the first stage is that local texture features dominate the shallow part of the network.
(iii) We achieve the best results when adding dual attention in all four stages.

\noindent\textbf{Dual attention layout.}~~
We conduct experiments on the layout of our dual attention. There are three options with similar computations: (i) window attention first; (ii) channel attention first; and (iii) two types of attention are paralleled arranged. The comparison is shown in Table~\ref{tab:dual_att_arch}. We can see that the three strategies achieve similar performance, with `window attention first' slightly better.

More ablative experiments can be found in Appendix.



\section{Conclusion}
This work introduces the dual attention mechanism, containing spatial window attention and channel group attention, to capture global contexts while maintaining computational efficiency. 
We show that these two self-attentions complement each other: 
(i) since each channel token contains an abstract representation of the entire image, the channel attention naturally captures global interactions and representations by taking all spatial positions into account when computing attention scores between channels;
(ii) spatial attention refines the local representations by performing fine-grained interactions across spatial locations, which in turn helps the global information modeling in channel attention.
We further visualize how our channel group attention captures global interactions and demonstrate its effectiveness in various benchmarks.


\appendix
\section{Appendix}

\begin{table}[t]
\begin{center}
    \caption{Model configurations for our \model. We introduce three configurations \model-Tiny, \model-Small, and \model-Base with different model capacities. The size of the input image is set to $224 \times 224$.}
\setlength{\tabcolsep}{2.1pt}
\renewcommand\arraystretch{1.3}
\footnotesize
    \centering
    \resizebox{1\linewidth}{!}{
    \begin{tabular}{l|c|c|c|c|c}
    \toprule
            &  Output Size & Layer Name & \model-Tiny & \model-Small & \model-Base \\
            \midrule
    \multirow{4}{*}{stage 1}  & $56\times56$    & Patch Embedding  &  $\text{kernel}~7, \text{stride}~4, \text{pad}~3, C^1=96$ 
    & $\text{kernel}~7, \text{stride}~4, \text{pad}~3, C^1=96$ 
    & $\text{kernel}~7, \text{stride}~4, \text{pad}~3, C^1=128$ \\
    \cmidrule{2-6}
                              & $56 \times 56$  &  \makecell{Dual \\ Transformer \\ Block} & \multicolumn{1}{c}{$\left[\begin{array}{c}
                                   \text{win. sz.}~7\times7, P_w=49 \\
                                   N_h^1=N_g^1=3 \\
                                   C_h^1=C_g^1=32
                              \end{array}\right] \times 1$} &
\multicolumn{1}{c}{$\left[\begin{array}{c}
                                   \text{win. sz.}~7\times7, P_w=49 \\
                                   N_h^1=N_g^1=3 \\
                                   C_h^1=C_g^1=32
                              \end{array}\right] \times 1$} &
\multicolumn{1}{c}{$\left[\begin{array}{c}
                                   \text{win. sz.}~7\times7, P_w=49 \\
                                   N_h^1=N_g^1=4 \\
                                   C_h^1=C_g^1=32
                              \end{array}\right] \times 1$}
                              \\
        \midrule
    \multirow{4}{*}{stage 2} & $ 28\times 28$ & Patch Embedding & 
    $\text{kernel}~2, \text{stride}~2, \text{pad}~0, C^2=192$ 
    & $\text{kernel}~2, \text{stride}~2, \text{pad}~0, C^2=192$ 
    & $\text{kernel}~2, \text{stride}~2, \text{pad}~0, C^2=256$ \\
    \cmidrule{2-6}
            & $ 28\times 28$ & \makecell{Dual \\ Transformer \\ Block} & \multicolumn{1}{c}{$\left[\begin{array}{c}
                                   \text{win. sz.}~7\times7, P_w=49 \\
                                   N_h^2=N_g^2=6 \\
                                   C_h^2=C_g^2=32
                              \end{array}\right] \times 1$} & 
\multicolumn{1}{c}{$\left[\begin{array}{c}
                                   \text{win. sz.}~7\times7, P_w=49 \\
                                   N_h^2=N_g^2=6 \\
                                   C_h^2=C_g^2=32
                              \end{array}\right] \times 1$} &
\multicolumn{1}{c}{$\left[\begin{array}{c}
                                   \text{win. sz.}~7\times7, P_w=49 \\
                                   N_h^2=N_g^2=8 \\
                                   C_h^2=C_g^2=32
                              \end{array}\right] \times 1$}                              \\
        \midrule
    \multirow{4}{*}{stage 3} & $ 14\times 14$ & Patch Embedding &
    $\text{kernel}~2, \text{stride}~2, \text{pad}~0, C^3=384$ 
    & $\text{kernel}~2, \text{stride}~2, \text{pad}~0, C^3=384$ 
    & $\text{kernel}~2, \text{stride}~2, \text{pad}~0, C^3=512$ \\
        \cmidrule{2-6}
            & $ 14\times 14$ & \makecell{Dual \\ Transformer \\ Block} & 
\multicolumn{1}{c}{$\left[\begin{array}{c}
                                   \text{win. sz.}~7\times7, P_w=49 \\
                                   N_h^3=N_g^3=12 \\
                                   C_h^3=C_g^3=32
                              \end{array}\right] \times 3$} & \multicolumn{1}{c}{$\left[\begin{array}{c}
                                   \text{win. sz.}~7\times7, P_w=49 \\
                                   N_h^3=N_g^3=12 \\
                                   C_h^3=C_g^3=32
                              \end{array}\right] \times 9$} & 
                 \multicolumn{1}{c}{$\left[\begin{array}{c}
                                   \text{win. sz.}~7\times7, P_w=49 \\
                                   N_h^3=N_g^3=16 \\
                                   C_h^3=C_g^3=32
                              \end{array}\right] \times 9$}\\
        \midrule
    \multirow{4}{*}{stage 4} & $ 7\times 7$ & Patch Embedding &
    $\text{kernel}~2, \text{stride}~2, \text{pad}~0, C^4=768$ 
    & $\text{kernel}~2, \text{stride}~2, \text{pad}~0, C^4=768$ 
    & $\text{kernel}~2, \text{stride}~2, \text{pad}~0, C^4=1024$ \\
    \cmidrule{2-6}
            & $ 7\times 7$ & \makecell{Dual \\ Transformer \\ Block} & \multicolumn{1}{c}{$\left[\begin{array}{c}
                                   \text{win. sz.}~7\times7, P_w=49 \\
                                   N_h^4=N_g^4=24 \\
                                   C_h^4=C_g^4=32
                              \end{array}\right] \times 1$} &
\multicolumn{1}{c}{$\left[\begin{array}{c}
                                   \text{win. sz.}~7\times7, P_w=49 \\
                                   N_h^4=N_g^4=24 \\
                                   C_h^4=C_g^4=32
                              \end{array}\right] \times 1$} &
\multicolumn{1}{c}{$\left[\begin{array}{c}
                                   \text{win. sz.}~7\times7, P_w=49 \\
                                   N_h^4=N_g^4=32 \\
                                   C_h^4=C_g^4=32
                              \end{array}\right] \times 1$}                              \\
            \bottomrule
    \end{tabular}}
    \vspace{-12pt}
    \label{tab:model_config}
\end{center}
\end{table}

\subsection{Details of Model Configuration}
In this work, we follow the design strategy suggested by previous works~\cite{liu2021swin,yang2021focal}.
We divide the entire architecture into four stages, where a patch embedding layer is inserted at the beginning of each stage.
Here, our patch embedding layer is implemented by stride convolution. The convolutional kernels and stride values of our four patch embedding layers are $\{7, 2, 2, 2\}$ and $\{4, 2, 2, 2\}$, respectively.
Note the large kernel in the first layer introduces almost no additional calculations as the number of input channels is only $3$. For the rest kernel values, we use 2 to perform non-overlapping patch merging.
We stack our dual attention blocks in each stage with the resolution and feature dimension kept the same.
These stages jointly produce a hierarchical representation, with the same feature map resolutions as those of typical convolutional networks, e.g., VGG~\cite{simonyan2014very} and ResNet~\cite{he2016deep}.
As a result, the proposed architecture can conveniently replace the backbone networks in existing methods for various vision tasks.
All model training is accelerated by NVIDIA Apex Automatic Mixed Precision (AMP).

Detailed model configurations of our tiny, small, and base models are shown in Table~\ref{tab:model_config}. 
Specifically, take an image with $H \times W$, a $C^1$-dimensional feature with a resolution of $\frac{H}{4} \times \frac{W}{4}$ is obtained after the first patch embedding layer. And its resolution is further reduced into $\frac{H}{8} \times \frac{W}{8}$, $\frac{H}{16} \times \frac{W}{16}$, and $\frac{H}{32} \times \frac{W}{32}$ with the feature dimension increasing to $C^2$, $C^3$, and $C^4$ after the other three patch embedding layer, respectively.
For simplicity, we set the window size of $7\times 7$ thus $P_w=49$ for all models. We also set the number of channels per group $C_g=32$ and the number of channels per head $C_h=32$ for all blocks of our three models.
For \model-tiny and \model-small, we set the number of heads/groups $N_h = N_g = \{3, 6, 12, 24\}$ for four stages, respectively; and we set the number of heads/groups $N_h = N_g = \{4, 8, 16, 32\}$ for four stages in \model-base.
Also, we set the number of dual attention blocks $\{1, 1, 3, 1\}$ for our tiny model and $\{1, 1, 9, 1\}$ for the small and base models.

For the models without FFN, we simply change the number of dual attention blocks to keep the total computation costs similar, though we believe there should be a better configuration specifically for them. We set the number of dual attention blocks $\{2, 2, 11, 2\}$ for our tiny model (without FFN) and $\{2, 2, 28, 2\}$ for the small and base models (without FFN).

When more training data is involved, we further scale up DaViT to large, huge, and giant sizes to validate the scaling ability of the proposed architecture for image classification with larger input resolutions.
Considering larger image resolutions and model sizes used, we suspect that the dot products of self-attention grow large in magnitude in this case, as in~\cite{vaswani2017attention}. To counteract this effect, we scale the dot products in our channel attention by $\frac{1}{\sqrt{P}}$.
We set $C^1 = \{192, 256, 384\}$, $N_h^1 = N_g^1 = \{6, 8, 12\}$, and the number of dual attention blocks of the third stage as $\{9, 9, 12\}$ respectively for large, huge, and giant models.

\begin{table}[t]
\footnotesize
\centering
\setlength{\tabcolsep}{2pt}
\renewcommand\arraystretch{1.2}
    \caption{Comparison on ImageNet-1K by replacing half of DeiT~\cite{touvron2021training} attention blocks with our channel attention block. Performance improvement over DeiT is highlighted in blue font.}
\vspace{-4pt}
\resizebox{0.7\linewidth}{!}{
    \begin{tabular}{l|ccc}
    \shline
    Model & \#Params (M) & FLOPs (G) & Top-1 (\%) \\
    \shline
    DeiT-Tiny~\cite{touvron2021training} & 5.7 & 1.2 & 72.2 \\
    \rowcolor{Gray}
    DeiT-Tiny~\cite{touvron2021training} + Channel Attention  & 5.7 & 1.2 &  \textbf{74.9}~\scriptsize{\textcolor{blue}{\bf(+2.7)}} \\
    \hline
    DeiT-Small~\cite{touvron2021training}  & 22.1 & 4.5 & 79.8 \\
    \rowcolor{Gray}
    DeiT-Small~\cite{touvron2021training} + Channel Attention & 22.1 & 4.5 & \textbf{81.2}~\scriptsize{\textcolor{blue}{\bf(+1.4)}} \\
    \hline
    DeiT-Base~\cite{touvron2021training}  & 86.7 & 17.4 & 81.8 \\
    \rowcolor{Gray}
    DeiT-Base~\cite{touvron2021training} + Channel Attention  & 86.7 & 17.4 & \textbf{82.3}~\scriptsize{\textcolor{blue}{\bf(+0.5)}} \\
     \shline
    \end{tabular}
    }
    \label{tab:deit}
\vspace{-4pt}
\end{table}

\subsection{Channel Attention on Vanilla ViT}
 We apply our channel group attention on the vanilla DeiT~\cite{touvron2021training} to show the generalizability and effectiveness of our dual attention mechanism. We alternatively arrange the vanilla patch-level self-attention and our channel group self-attention.
 We set the number of groups and channels of our channel-wise group attention the same as the number of heads and channels of self-attention in DeiT, to make the number of parameters and FLOPs comparable with the vanilla DeiT. From Table~\ref{tab:deit} we observe substantial gains of our dual attention across all model sizes, \eg, 2.7\% over the tiny model and 0.5\% even compared to the base model.
 The result shows that our channel-wise attention can be combined with window attention and patch-level global attention, improving the performance of both spatial-wise self-attentions.

\begin{table}[t]
\footnotesize
\centering
\caption{Comparison of Transformers without FFNs on ImageNet-1K. FFN is removed and more attention layers are added to match the computational cost.}
\vspace{-4pt}
\setlength{\tabcolsep}{5pt}
\renewcommand\arraystretch{1.2}
\resizebox{0.7\linewidth}{!}{
    \begin{tabular}{l|ccc}
    \shline
    Model (w/o FFN) & \#Params & FLOPs & Top-1 (\%) \\
    \shline
    Window Attention-Tiny (w/o FFN) & 25.8 & 4.6 & 79.1 \\
    Channel Attention-Tiny (w/o FFN)  & 25.8 & 4.5 & 79.3 \\
    Dual Attention-Tiny (w/o FFN) & 25.8 & 4.5 & 80.8 \\
    \hline
    Dual Attention-Small (w/o FFN) & 46.3 & 8.7 & 81.9 \\
    Dual Attention-Base (w/o FFN) & 81.6 & 15.2 & 82.5 \\
     \shline
    \end{tabular}
    }
    \label{tab:without_ffn}
\end{table}

\subsection{Transformer without FFNs}
FFN has been a default component of Transformers with little research on it. However, it dominates the number of FLOPs and model parameters of our \model. Considering our dual attention has both channel-wise and spatial-wise interactions, we conduct an initial exploration to show the potential of the pure-attention structure without FFNs.
We remove FFNs and add more dual attention blocks to match the computational costs.

From Table~\ref{tab:without_ffn} we see that: pure dual attention without FFNs achieves 1.5\% and 1.7\% better Top-1 accuracy than pure window attention and channel attention, respectively, showing the effectiveness of dual attention that has both channel-wise and spatial-wise interactions.
The model without FFN shows comparable and even better performance with models like PVT~\cite{wang2021pyramid} and DeiT~\cite{touvron2021training}, but still inferior to recent SoTAs like Swin~\cite{liu2021swin}, Focal~\cite{yang2021focal}, and our full \model.

\begin{table}[t]
\setlength{\tabcolsep}{8pt}
\renewcommand\arraystretch{1.2}
\centering
\caption{Impact of the change of model depth. We gradually reduce the number of transformer layers at the third stage from the original 3 (6 in Swin~\cite{liu2021swin} and Focal~\cite{yang2021focal}) to 2 (4) and further 1 (2).}
\vspace{-4pt}
\footnotesize
\resizebox{0.7\linewidth}{!}{
  \begin{tabular}{cc|ccc}
    \shline
    Depths & Model & \#Params. (M) &  FLOPs (G) & Top-1~(\%) \\
    \shline
    \multirow{2}{*}{2-2-2-2}   & Swin~\cite{liu2021swin}  & 21.2 & 3.1 & 78.7  \\
                                 & Focal~\cite{yang2021focal}  & 21.7 & 3.4 & 79.9 \\
      \rowcolor{Gray}        1-1-1-1               & \model (ours) & 21.2 & 3.1 & \textbf{80.2} \\ 
      \hline
    
    \multirow{2}{*}{2-2-4-2}   & Swin~\cite{liu2021swin} & 24.7 & 3.8 & 80.2 \\
                                 & Focal~\cite{yang2021focal} & 25.4 & 4.1 & 81.4 \\  
      \rowcolor{Gray}       1-1-2-1               & \model (ours) & 24.7 & 3.8 & \textbf{81.8} \\
      \hline
    \multirow{2}{*}{2-2-6-2}   & Swin~\cite{liu2021swin} & 28.3 & 4.5 & 81.2  \\
                                 & Focal~\cite{yang2021focal} & 29.1 & 4.9 & 82.2 \\
      \rowcolor{Gray}       1-1-3-1               & \model (ours) & 28.3 & 4.5 & \textbf{82.8} \\
                              \shline
  \end{tabular}
  }
  \label{tab:performance_with_depth}
\end{table}

\subsection{Model Capacity against Model Depth}
Our \model contains both spatial-wise and channel-wise attention, and both local and global interactions in each dual attention block. We conduct experiments to show whether it needs less number of layers to obtain similar modeling capacity as previous works~\cite{liu2021swin,yang2021focal}.
We gradually reduce the number of transformer layers at the third stage from original 3 (6 in Swin~\cite{liu2021swin} and Focal~\cite{yang2021focal}) to 2 (4 in Swin~\cite{liu2021swin} and Focal~\cite{yang2021focal}) and further 1 (2 in Swin~\cite{liu2021swin} and Focal~\cite{yang2021focal}).
From Table~\ref{tab:performance_with_depth}, we find our model outperforms existing SoTA models consistently with the same depth. Moreover, using two fewer layers, our model achieves comparable and even better performance to Swin Transformer. Also, with fewer computational costs, our model outperforms Focal Transformer by a large margin.

\begin{figure}[t]
\begin{minipage}{0.5\linewidth}
\begin{table}[H]
\centering
\caption{Throughput of Swin and DaViT.}
\setlength{\tabcolsep}{4pt}
\renewcommand\arraystretch{1.05}
\begin{tabular}{lr}
\shline
Model & Throughput (samples/s) \\
\shline
Swin-T  & 1024                                                             \\
DaViT-T & 1059                                                             \\ \hline
Swin-S  & 655                                                             \\
DaViT-S & 685                                                             \\ \hline
Swin-B  & 496                                                             \\
DaViT-B & 523                                                             \\ 
\shline
\end{tabular}
\label{tab:throughput}
\end{table}
\end{minipage}
\begin{minipage}{0.49\linewidth}
\begin{table}[H]
\centering
\caption{Comparisons to the channel-wise attention used in CNNs, by replacing our channel group attention with SENet~\cite{hu2018squeeze} and ECANet~\cite{wang2020eca}, respectively.}
\setlength{\tabcolsep}{4pt}
\renewcommand\arraystretch{1.1}
\begin{tabular}{lr}
\shline
Method & Top-1 (\%) \\
\shline
SE Block~\cite{hu2018squeeze}  & 81.2        \\
ECA Block~\cite{wang2020eca} & 81.2  \\ 
Channel Self-Attention &  82.8     \\ 
\shline
\end{tabular}
\label{tab:attention}
\end{table}
\end{minipage}
\end{figure}

\subsection{Throughput Analysis}
In addition to the main criteria of computational cost in the main paper, \ie, FLOPs and \#parameters, we report the real-time inference speed/throughput against Swin Transformer~\cite{liu2021swin} to show the efficiency of our work.
Compared to the strong baseline Swin, our model does have advantages in real-time throughput as the cleaner structure and high efficiency of the group channel attention. For example, we remove the shifted window partition, which depends on \texttt{torch.roll()} to perform cyclic shift and its reverse on features. This operation is not fully optimized by popular inference frameworks. Table~\ref{tab:throughput} demonstrates the comparison between Swin and DaViT. Nvidia V100 GPU is utilized for the benchmark, and the image resolution is $224\times224$. It shows that DaViT consistently outperforms Swin across different model sizes in efficiency.

\subsection{Comparisons with SE and ECA Blocks}
To make quantitative comparisons with traditional channel-wise attentions blocks, we did experiments by replacing the channel self-attention in our tiny model with SE block~\cite{hu2018squeeze} and ECA block~\cite{wang2020eca}.
The results in Table~\ref{tab:attention} show that our channel group self-attention is more powerful by performing dynamic feature fusion across global tokens (different global views of the entire image) in transformers. It shows superior performance (1.6\%) than both of the two variants using SE~\cite{hu2018squeeze} and ECA~\cite{wang2020eca} blocks, respectively.

\bibliographystyle{splncs04}
\bibliography{egbib}
\end{document}